\documentclass[journal]{IEEEtran}

\IEEEoverridecommandlockouts
\usepackage{cite}
\usepackage{lipsum}
\usepackage{multirow}
\usepackage{amsmath,amssymb,amsfonts}
\usepackage{graphicx}
\usepackage{textcomp}
\usepackage{xcolor}
\usepackage{makecell}
\usepackage{array}
\usepackage{threeparttable}
\usepackage{algorithm}
\usepackage{algorithmic}
\usepackage{amsthm}
\usepackage{booktabs}
\usepackage{bm}
\usepackage{tabu}
\usepackage{url}

\newcommand{\etal}{\textit{et al}.}
\newcommand{\ie}{\textit{i}.\textit{e}.}
\newcommand{\eg}{\textit{e}.\textit{g}.}
\usepackage{gensymb}
\usepackage{arydshln}
\usepackage{amsmath,amsfonts,amssymb}
\usepackage{algorithmic}
\usepackage{algorithm}
\usepackage{array}

\usepackage{subfigure}
\usepackage{graphicx}
\usepackage{textcomp}
\usepackage{stfloats}
\usepackage{url}
\usepackage{verbatim}
\usepackage{color}
\usepackage{multirow}
\usepackage{booktabs}

\usepackage{bm}

\usepackage{xcolor}
\usepackage{hyperref}
\usepackage{makecell}

\makeatletter
\newcommand*{\rom}[1]{\expandafter\@slowromancap\romannumeral #1@}
\makeatother
\usepackage{makecell}
\usepackage{threeparttable}

\newlength\savedwidth
\newcommand\whline{\noalign{\global\savedwidth\arrayrulewidth\global\arrayrulewidth 1.0pt}
\hline\noalign{\global\arrayrulewidth\savedwidth}}

\begin{document}
\title{Perceptual Quality Assessment of Virtual Reality Videos in the Wild}

\author{Wen~Wen,
        Mu~Li,
        Yiru~Yao,
        Xiangjie~Sui,
        Yabin~Zhang,
        Long~Lan,
        Yuming~Fang,~\IEEEmembership{Senior Member,~IEEE,}
        and Kede~Ma,~\IEEEmembership{Senior Member,~IEEE}

\thanks{This work was supported in part by National Natural Science Foundation of China (NSFC) under Grants 62102339 and 62132006, Shenzhen Science and Technology Program, PR China under Grant RCBS20221008093121052 and GXWD20220811170130002, and the Hong Kong RGC Early Career Scheme (9048212). \textit{(Corresponding Author: Mu Li)}}
\thanks{Wen Wen and Kede Ma are with the Department of Computer Science, City University of Hong Kong, Kowloon, Hong Kong (e-mail: wwen29-c@my.cityu.edu.hk, kede.ma@cityu.edu.hk).}
\thanks{Mu Li is with the School of Computer Science and Technology, Harbin Institute of Technology, Shenzhen 518055, Guangdong, China (e-mail: limuhit@gmail.com).}
\thanks{Yiru Yao, Xiangjie Sui, and Yuming Fang are with the School of Information Management, Jiangxi University of Finance and Economics, Nanchang 330032, Jiangxi, China (e-mail: yaoyiru1998@foxmail.com, suixiangjie2017@163.com, fa0001ng@e.ntu.edu.sg).}
\thanks{Yabin Zhang is with the Multimedia Lab, Bytedance Inc. (e-mail: zhan0398@e.ntu.edu.sg).}
\thanks{Long Lan is with the Institute for Quantum Information \& State Key Laboratory of High Performance Computing, College of Computer Science and Technology, National University of Defense Technology, Changsha 410073, China (e-mail: long.lan@nudt.edu.cn).}
}

\markboth{IEEE TRANSACTIONS ON CIRCUITS AND SYSTEMS FOR VIDEO TECHNOLOGY}
{Shell \MakeLowercase{\textit{et al.}}: Bare Demo of IEEEtran.cls for IEEE Journals}

\IEEEpubid{\begin{minipage}{\textwidth}\ \centering
Copyright © 2024 IEEE. Personal use of this material is permitted. \\
However, permission to use this material for any other purposes must be obtained from the IEEE by sending an email to pubs-permissions@ieee.org.
\end{minipage}}

\maketitle

\begin{abstract}
\fontdimen2\font=0.5ex
Investigating how people perceive virtual reality (VR) videos in the wild (\ie, those captured by everyday users) is a crucial and challenging task in VR-related applications due to  complex \textit{authentic} distortions localized in space and time. Existing panoramic video databases only consider synthetic distortions, assume fixed viewing conditions, and are limited in size. To overcome these shortcomings, we construct the VR Video Quality in the Wild (VRVQW) database, containing $502$ user-generated videos with diverse content and distortion characteristics. Based on VRVQW, we conduct a formal psychophysical experiment to record the scanpaths and perceived quality scores from $139$ participants under two different viewing conditions. We provide a thorough statistical analysis of the recorded data, observing significant impact of viewing conditions on both human scanpaths and perceived quality. Moreover, we develop an objective quality assessment model for VR videos based on pseudocylindrical representation and convolution. Results on the proposed VRVQW show that our method is superior to existing video quality assessment models. 
We have made the database and code available at \url{https://github.com/limuhit/VR-Video-Quality-in-the-Wild}.
\end{abstract}


\IEEEpeerreviewmaketitle

\section{Introduction}
\fontdimen2\font=0.5ex
\IEEEPARstart{A}{s} virtual reality (VR) acquisition and display systems become widely accessible, people are getting used to capturing, editing, and interacting with VR content, which is evidenced by the accelerated proliferation of panoramic~videos\footnote{In this paper, we use the terms ``panoramic'', ``VR'', ``360\degree'', ``omnidirectional'', and ``spherical'', interchangeably.} uploaded to popular video sharing and social media platforms (\eg, Bilibili and YouTube). A practical issue arising from panoramic videos in the wild is that they are often born with complex visual artifacts (\ie, the so-called \textit{authentic} distortions) due to scene complexity, lens imperfection, sensor limitation, non-professional shooting, and stitching inaccuracy. The acquired videos may subsequently undergo several stages of processing, including compression, editing, transmission, and transcoding, leading to additional video impairments~\cite{2020VRDistortions}. Understanding how people perceive 360\degree~video distortions in virtual environments is central to many VR-enabled video applications. \IEEEpubidadjcol

\begin{figure*}[t]
    \begin{center}
        \includegraphics[width=\linewidth]{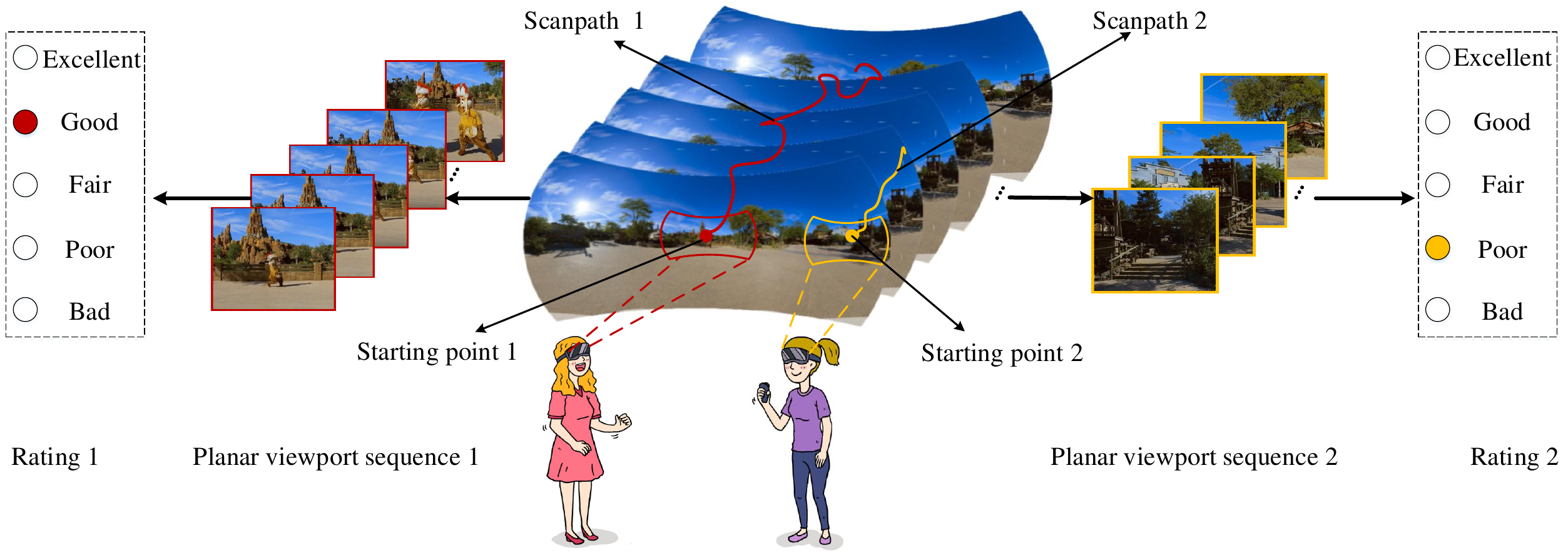}
        \caption{Illustration of how people explore VR videos in the proposed VRVQW database. Under varying viewing conditions (\eg, starting points and exploration times), users may exhibit different viewing behaviors in the form of scanpaths, leading to different portions of the video being explored. As user-generated VR videos often come with localized authentic distortions, the perceived quality may vary with user viewing behaviors constrained by viewing conditions. 
        Therefore, the incorporation of viewing conditions would be the key to the success of computational quality prediction of user-generated VR videos.
        }
        \label{fig:teaser} 
    \end{center}
\end{figure*}

Different from a planar video, a panoramic video, by its name, records/generates the scene of interest by capturing/tracing light from all directions at possibly varying viewpoints through time. This gives rise to a $360\degree \times 180\degree$ spherical field of view (FoV) at any time instance. With the help of a head-mounted display (HMD), users can freely explore the virtual scene using  head and gaze movements as if they were in the real world. Such immerse and interactive viewing experience poses a great challenge to existing quality assessment methods for planar videos \cite{Wang2004VQA,Wang2012VQA,Xu2020_C3DVQA,Li2019VSFA,Korhonen2019_TLVQA,Ying2021Patch_up,Franz2021_MLSP-VQA} when predicting 360\degree~video quality. Although several subjective quality studies \cite{Singla2017OVD,Curcio2017OVD,Tran2017OVD,IVQAD2017,Zhang2018_subjective,ZhangY2018OVD,Lopes2018subjective,li2018_subjective,Meng2021OVD} on panoramic videos have been conducted, they may suffer from three limitations. First, most of the resulting databases contain \textit{synthetic} distortions only, with compression artifacts being the most representative. This is an oversimplification of the real-world situation, where user-generated 360{\degree} videos may suffer from commingled \textit{authentic} distortions, often \textit{localized} spatiotemporally. The extensively studied compression artifacts may no longer  dominate the perceptual quality.  Second, the databases assume the viewing conditions such as the starting point and  exploration time to be fixed, which is overly restrictive when viewing virtual scenes with HMDs. Relaxing these constraints can lead to situations where the visible distortions of a 360{\degree} video are probably not perceived under some viewing conditions. Thus it is reasonable to rate the perceptual quality as high. Third, given a fixed human labeling budget, the number of unique reference 360{\degree} videos is determined by the number of synthetic distortion types and levels, which is limited to a few dozen (if not fewer). As such, these databases fail to sufficiently represent real-world videos with diverse content, distortion, and motion complexities.

Considering the aforementioned limitations of existing 360{\degree}~video databases,  several significant aspects related to understanding the perceptual quality of 360{\degree}~videos remain unexplored. For example, how consistent are human behaviors under the same viewing condition? How are human behaviors affected by viewing conditions? How does the perceived quality change with viewing conditions? Can we make effective computational quality predictions under different viewing conditions? In an attempt to answer these questions, we establish a large 360{\degree}~video database called the VR Video Quality in the Wild (VRVQW) database. VRVQW  comprises $502$ panoramic video sequences that cover a wide range of scenes, including cityscape, landscape, shows, sports, and computer-generated (CG) content. The videos also exhibit a wide range of complex authentic distortions, covering the full quality spectrum. To evaluate the subjective quality of a panoramic video as a function of viewing conditions, we invite subjects to watch the video from different starting points and time durations (see Fig.~\ref{fig:teaser}). A total of $40,268$ human opinion scores, along with scanpaths (as
viewing behaviors) from $139$ users, are recorded. We then provide an in-depth analysis of our data, investigating the impact of viewing conditions on viewing behaviors and perceived quality.

Moreover, we take initial steps toward blind video quality assessment (VQA) of VR content. To respect the spherical natural of VR videos while maintaining a manageable computational budget, we resort to the pseudocylindrical representation~\cite{li2021pseudocylindrical} in place of and as a generalization of the default equirectangular projection (ERP) format. On top of it, we design a lightweight convolutional neural network (CNN) for 360{\degree} VQA, in which the main operation---pseudocylindrical convolution---is fully compatible with the pseudocylindrical representation, and can be efficiently implemented by standard convolution. We explicitly incorporate two viewing conditions---the starting point and  exploration time---into the model design by rotating the initial pseudocylindrical representation and adjusting the sampling rate, respectively.
Experimental results show that our method effectively addresses the unique VQA challenges presented in the proposed VRVQW database, and performs favorably against existing methods. 

In summary, our main contributions are:
\begin{itemize}
    \item a first-of-its-kind 360{\degree} video database, VRVQW, with complex authentic distortions,
    \item a formal psychophysical experiment to record human scanpaths and quality scores under two viewing conditions, accompanied by thorough statistical analysis, and
    \item a lightweight blind VQA model specifically for VR videos, aware of viewing conditions.
\end{itemize}

\begin{table*}[t]
	\centering
	\renewcommand{\arraystretch}{1.25}
        \caption{Summary of VR VQA databases. ERP, RCMP, and TSP stand for the equirectangular projection, the reshaped cubemap projection, and the truncated square pyramid projection, respectively. The numbers in the ``\#videos'' column are in the form of ``\#reference videos / \#distorted videos''}
        \label{tab:summary}
        \resizebox{1\textwidth}{!}{
        \begin{tabular}{l|cccccccc}
\whline
Database           & Year & Projection  & \#videos  & \#subjects & Resolution                                                         & Duration (sec) & HM/EM data & Distortion type                                                                                                         \\ \hline
Singla \etal~\cite{Singla2017OVD}             & 2017 & ERP         & $6$ / $60$    &$30$     & \begin{tabular}[c]{@{}c@{}}$1,920\times1,080$ to\\  $3,840\times2,160$\end{tabular} & $10$       & HM         & H.265 compression, Downsampling                                                                                                      \\ \hline
Curcio \etal~\cite{Curcio2017OVD}            & 2017 & ERP         & $3$ / $24$     &$12$    & $3,840\times1,920$                                                           & $21$       & HM        & Tile-based H.265 compression                                                                                                       \\ \hline
Tran \etal~\cite{Tran2017OVD}               & 2017 & ERP         & $3$ / $60$     &$37$    & \begin{tabular}[c]{@{}c@{}}$1,440\times720$ to\\  $3,840\times1,920$\end{tabular}  & $30$       & N/A        & H.264 compression                                                                                                       \\ \hline
IVQAD2017\cite{IVQAD2017}  & 2017 & ERP         & $10$ / $150$  &$13$     & $4,096\times2,048$                                                           & $15$       & N/A        & MPEG-4 compression, Downsampling                                                                                           \\ \hline
BIT360~\cite{Zhang2018_subjective}   & 2017 & ERP         & $16$ / $384$    &$23$   & $4,096\times2,048$                                                           & $10$       & N/A        & \begin{tabular}[c]{@{}c@{}}H.264 compression H.265 compression, \\ VP9 compression, Simulated packet loss\end{tabular} \\ \hline
Zhang \etal~\cite{ZhangY2018OVD}  & 2018 & ERP         & $10$ / $50$    &$30$    & $3,600\times 1,800$                                                           & $10$       & N/A        & H.265 compression, Downsampling                                                                                            \\ \hline
Lopes \etal~\cite{Lopes2018subjective} & 2018 & ERP         & $6$ / $79$    &$37$     & \begin{tabular}[c]{@{}c@{}}$960\times480$ to \\ $7,680\times3,840$\end{tabular}   & $10$       & N/A        & H.265 compression, Downsampling                                                                                           \\ \hline
VQA-ODV \cite{li2018_subjective}   & 2018 & ERP, RCMP, TSP & $60$ / $540$   &$221$    & \begin{tabular}[c]{@{}c@{}}$3,840\times1,920$ to \\ $7,680\times3,840$\end{tabular} & $10$ to $23$ & HM + EM    & H.265 compression, Projection                                                                                             \\ \hline
VOD-VQA \cite{Meng2021OVD}            & 2021 & ERP         & $18$ / $774$    &$160$   & $3,840\times1,920$   & $10$       & N/A       & H.264 compression, Downsampling                                                                                           \\ \hline
VRVQW (Proposed)       & 2021 & ERP         & $-$ / $502$    & $139$   & \begin{tabular}[c]{@{}c@{}}$1,280 \times 720$ to \\ $5,120 \times 2,560$\end{tabular}  & $15$ & HM + EM   & Authentic distortion \\ 
\whline
\end{tabular}
}
\end{table*}

\section{Related Work}
In this section, we detail the current landscape of 360{\degree} VQA databases with mean opinion scores (MOSs). We next review objective VQA models that have either been adapted or specifically designed to evaluate VR content.

\subsection{Subjective Quality Assessment of Panoramic Videos}
Singla \etal~\cite{Singla2017OVD} constructed one of the first  databases to study the impact of H.265 compression and spatial resolution on 360{\degree} video quality. The database contains six reference videos and $60$ distorted videos at two resolutions and five bitrates. 
Curcio \etal~\cite{Curcio2017OVD} performed a subjective user study of 360{\degree} videos under the  tile-based streaming setting~\cite{Mario2017Streaming}. The visual stimuli consist of $24$ distorted videos at four quality levels and two resolutions, carefully selected to probe whether the background tile should be encoded with higher resolution or higher fidelity given the same bitrate budget. Tran \etal~\cite{Tran2017OVD} established a small database containing $60$ mobile distorted videos at five levels of H.264 compression and four resolutions. Duan \etal~\cite{IVQAD2017} studied the impact of MPEG-4 compression and spatial resolution on the perceptual quality of 360{\degree} videos. Zhang \etal~\cite{Zhang2018_subjective} proposed a large omnidirectional video dataset, including $16$ reference and $384$ distorted videos, covering H.264, H.265, VP9 compression, and simulated packet loss. They also proposed a standardized subjective procedure with improved efficiency. Zhang \etal~\cite{ZhangY2018OVD} conducted a comprehensive study on the interaction between subsampling and H.264 compression to panoramic video quality. They computed an optimal resolution, $3,600\times1,800$, for the HTC VIVE display.
Lopes \etal~\cite{Lopes2018subjective} studied the individual and combined effects of spatial resolution, frame rate, and H.265 compression to 360{\degree} videos. Li \etal~\cite{li2018_subjective} introduced the VQA-ODV dataset, consisting of $540$ impaired 360{\degree} videos from $60$ references using different levels of H.265 compression and map projections. VQA-ODV also includes head movement (HM) and eye movement (EM) data, along with an analysis of human behavior consistency. VOD-VQA~\cite{Meng2021OVD} is currently the largest panoramic video database, which includes $18$ reference and  $774$ distorted videos with different compression levels, spatial resolutions, and frame rates.

We present a summary of existing 360{\degree} video databases in Table~\ref{tab:summary}, where we observe that these databases primarily consist of \textit{synthetic} distortions, assuming the availability of original undistorted videos for database construction. In contrast, we focus on user-generated panoramic videos in the wild, many of which suffer from \textit{authentic} distortions during video acquisition. Moreover, it is not uncommon to see that these distortions are \textit{localized} in space and time, making viewing conditions indispensable for determining VR video quality~\cite{Sui2021}.

\subsection{Objective Quality Assessment of Panoramic Videos}
Existing objective quality models for panoramic content are mainly adapted from planar image quality assessment (IQA) and VQA methods, applied to one of three data formats: (projected) 2D plane, spherical surface, and (projected) rectilinear viewport. 

Methods in the planar domain~\cite{Sun2017_wspsnr,Vladyslav2016,Kim2019VRIQANET_OIQA} aim to compensate for the non-uniform sampling caused by the sphere-to-plane projection. When using ERP, planar I/VQA methods can be enhanced by latitude-dependent weighting. Another approach is to use the Craster parabolic projection to ensure uniform sampling density~\cite{Vladyslav2016}. Kim \etal~\cite{Kim2019VRIQANET_OIQA} explored an adversarial loss for learning patch-based quality estimators using content and position features. Li \etal~\cite{li2018_subjective} trained a CNN for panoramic VQA, leveraging the HM and EM data.
Methods in the spherical domain (\eg, S-PSNR~\cite{Yu2015SPSNR_OIQA} and S-SSIM~\cite{Chen2018SSIM_OIQA})  calculate and aggregate local quality estimates over the sphere. 
Yu \etal~\cite{Yu2015SPSNR_OIQA} incorporated importance weighting derived from the empirical  distributions of the HM and EM data.
Methods in the viewport domain focus on extracting viewports that are likely to be seen for quality computation. Xu \etal~\cite{Xu2021VGCN} used graph convolutional networks to model the spatial relations of extracted viewports, which, however, does not necessarily reflect the human viewing process. 
Li \etal~\cite{Li2019_VQA} proposed a two-stage approach, involving viewport proposal and quality assessment with spherical convolution~\cite{Taco2018_ShpericalCNN}. Recently, Sui \etal~\cite{Sui2021} suggested converting a panoramic image to planar videos by sampling sequences of rectilinear projections of viewports along users' scanpaths. This approach allows mature planar I/VQA methods to be directly applied.

\section{Subjective Quality Assessment of VR Videos in the Wild}
In this section, we summarize our effort toward creating the VRVQW database
that contains human perceptual data - MOSs and scanpaths by recording users' responses when watching 360{\degree} videos under different viewing conditions.

\subsection{Data Gathering}\label{subsec:dg}

\textit{1) Visual Stimuli:} The VRVQW database contains $502$ unique user-generated 360\degree~video sequences, with frame rates ranging from $20$ to $60$ frames per second (fps) and resolutions ranging from $1,280 \times 720$ to $5,120 \times 2,560$ pixels. 
All videos are collected from the Internet and carry Creative Commons licenses. Each video is cropped to a duration of about $15$ seconds and stored in the ERP format without further compression. VRVQW mainly includes 360{\degree} videos shot by a (nearly) static camera to reduce the probability of causing dizziness~\cite{Kim2019VRsickness}, which could potentially affect the reliability of the collected MOSs. We only select $32$ videos with significant camera motion, ensuring visual comfort through a posteriori questionnaire~\cite{Kennedy1993SSQ}. These moving-camera videos typically receive a considerable number of likes on video-sharing platforms, and provide users with a stronger sense of immersion and interaction.

The video selection process aim to encompass a range of scenes suitable for VR shooting, including \textit{Cityscape}, \textit{Landscape}, \textit{Shows}, \textit{Sports}, \textit{CG}, and \textit{Others}. \textit{Cityscape} contains different places of interest around the world like the Roman Colosseum and other renowned historical sites. \textit{Landscape} includes beautiful natural scenes, such as waterfalls, mountains, volcanoes, etc. \textit{Shows} represent different forms of entertainment, including band performance, living theatre, and street improvisation. \textit{Sports} gather various sporting events, \eg, car racing, skiing, and riding. \textit{CG} is a collection of rendered videos by mature computer graphics techniques.  Finally, the \textit{Others} category is reserved for scenes that do not belong to the previous five classes.

\begin{figure}[t]
    \begin{centering}
        \includegraphics[scale = 0.35]{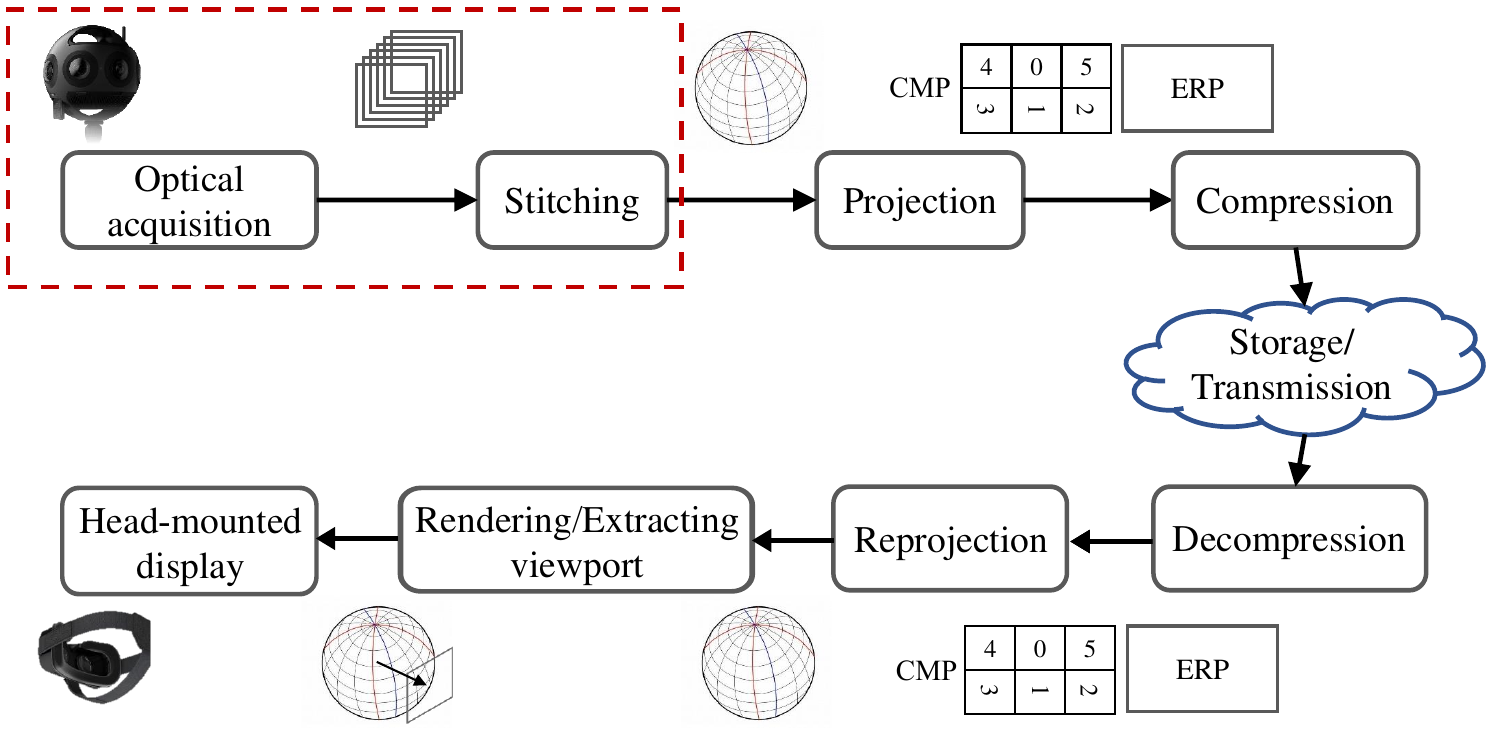}
        \caption{360\degree~video processing pipeline, from optical acquisition to content consumption via an HMD. The optical acquisition and stitching are two main steps for 360{\degree} video creation, where authentic distortions arise.}
        \label{fig:VR_video_process} 
    \end{centering}
\end{figure}

\begin{figure*}[t]
    \centering
        \subfigure[Under-exposure]{\includegraphics[width=0.225\linewidth]{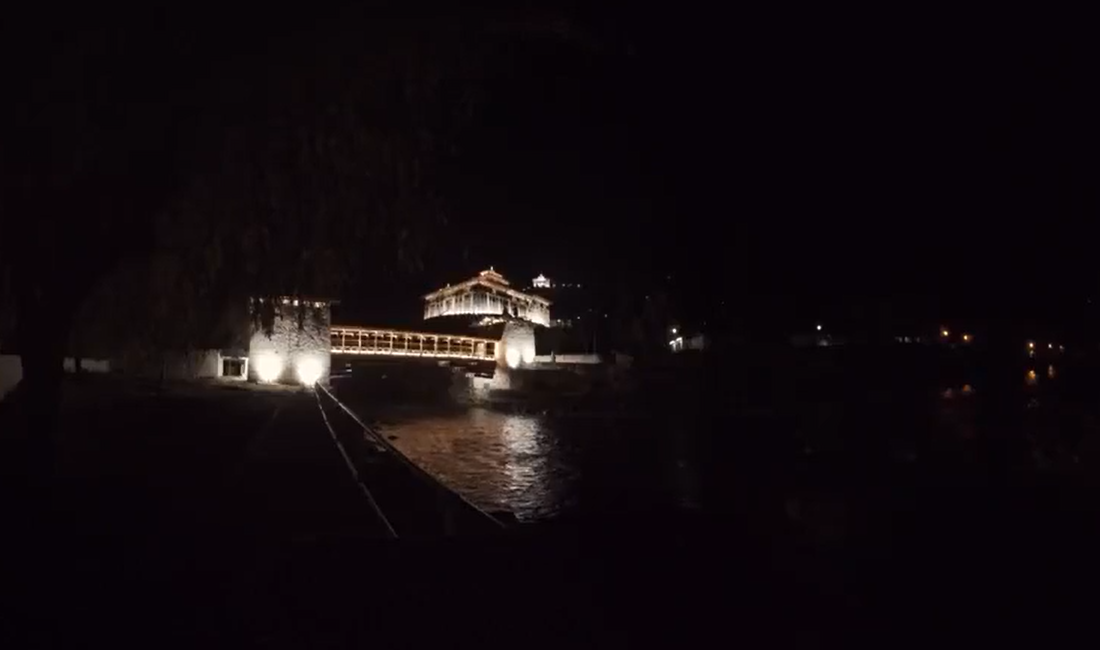}}\hskip.8em
        \subfigure[Over-exposure]{\includegraphics[width=0.225\linewidth]{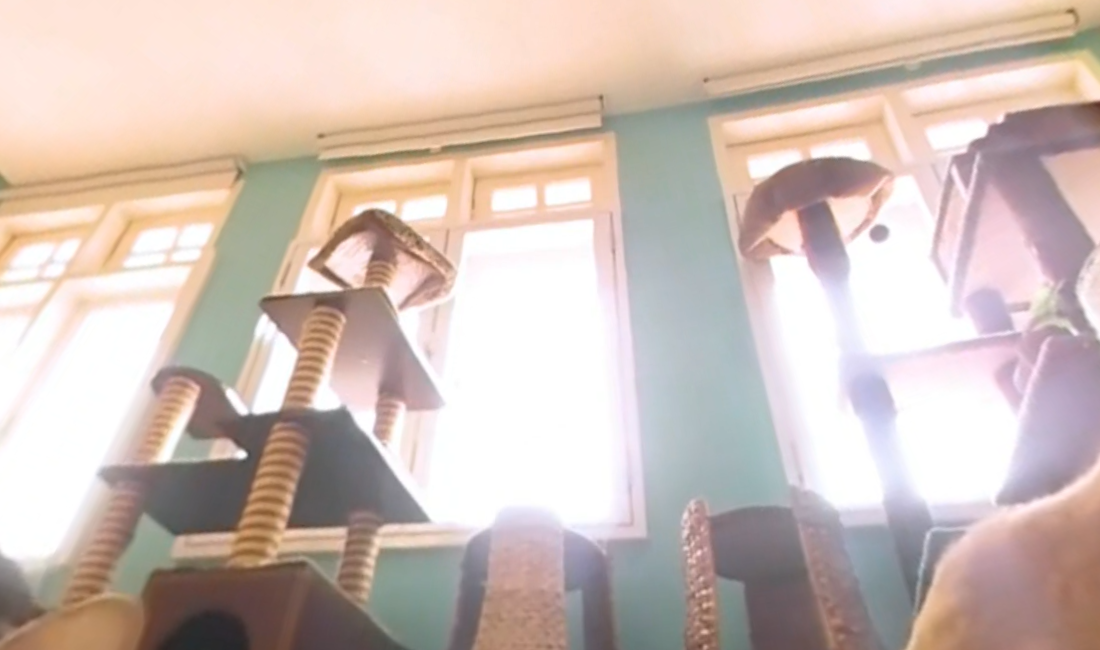}}\hskip.8em
        \subfigure[Motion blurring]{\includegraphics[width=0.225\linewidth]{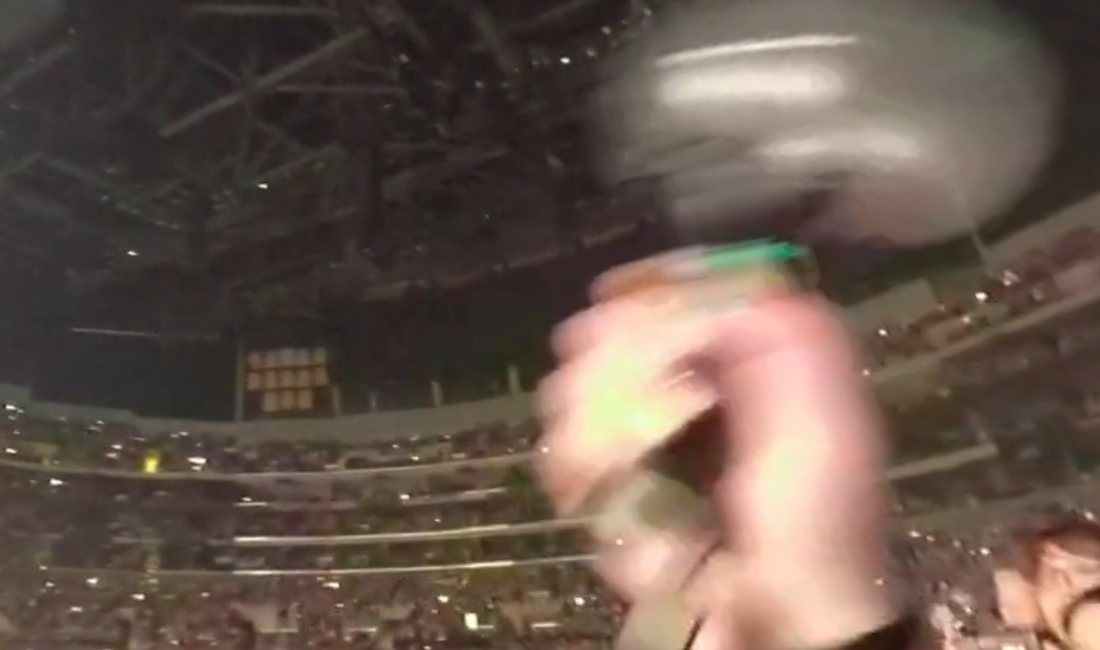}}\hskip.8em
        \subfigure[Abrupt luminance change]{\includegraphics[width=0.225\linewidth]{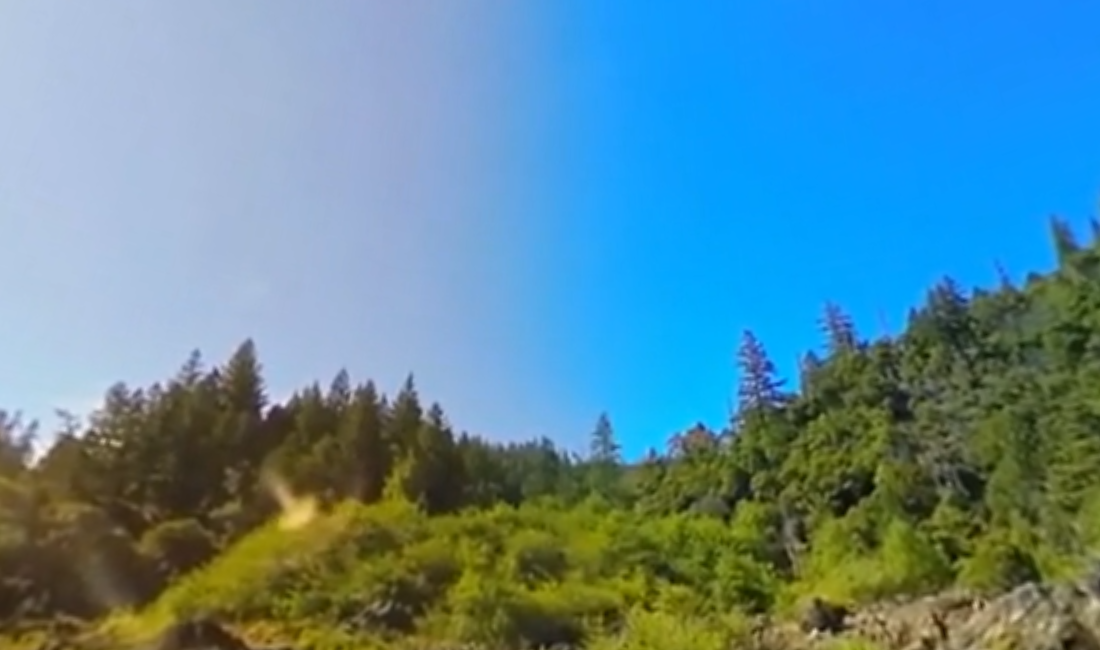}}\\
        \subfigure[Abrupt structure change]{\includegraphics[width=0.225\linewidth]{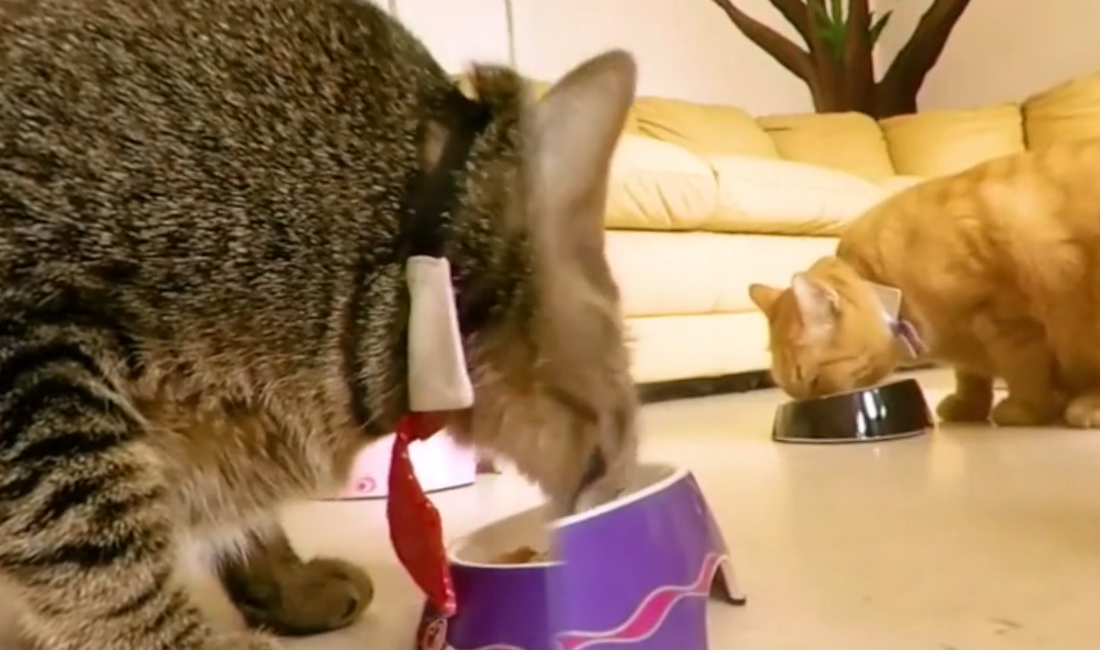}}\hskip.8em
        \subfigure[Object with missing parts]{\includegraphics[width=0.225\linewidth]{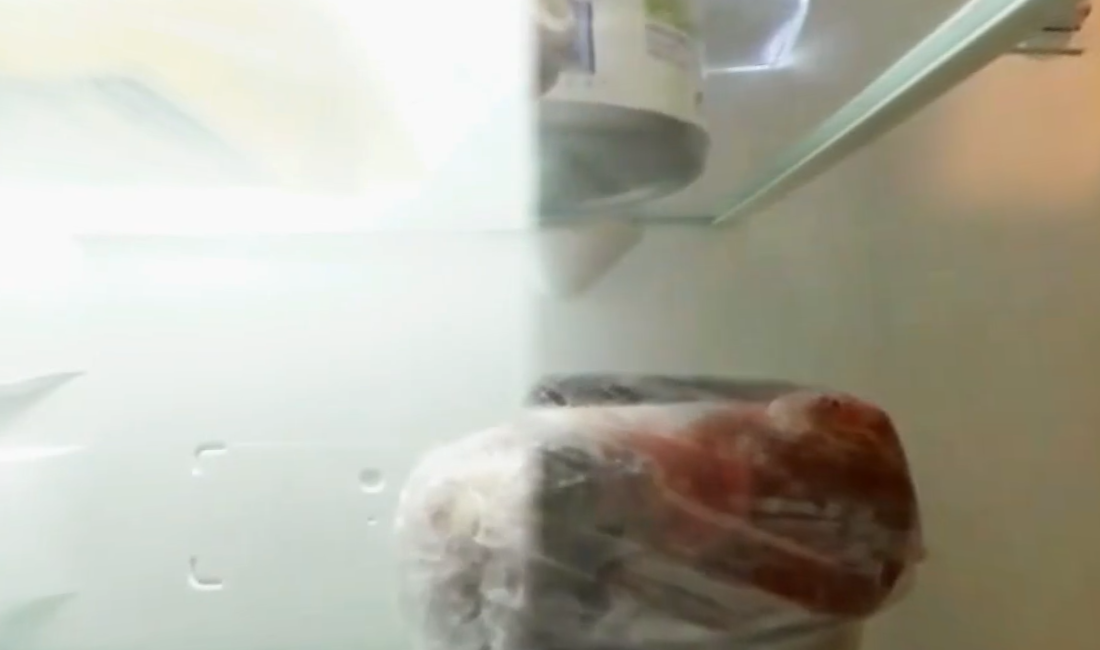}}\hskip.8em
        \subfigure[Ghosting]{\includegraphics[width=0.225\linewidth]{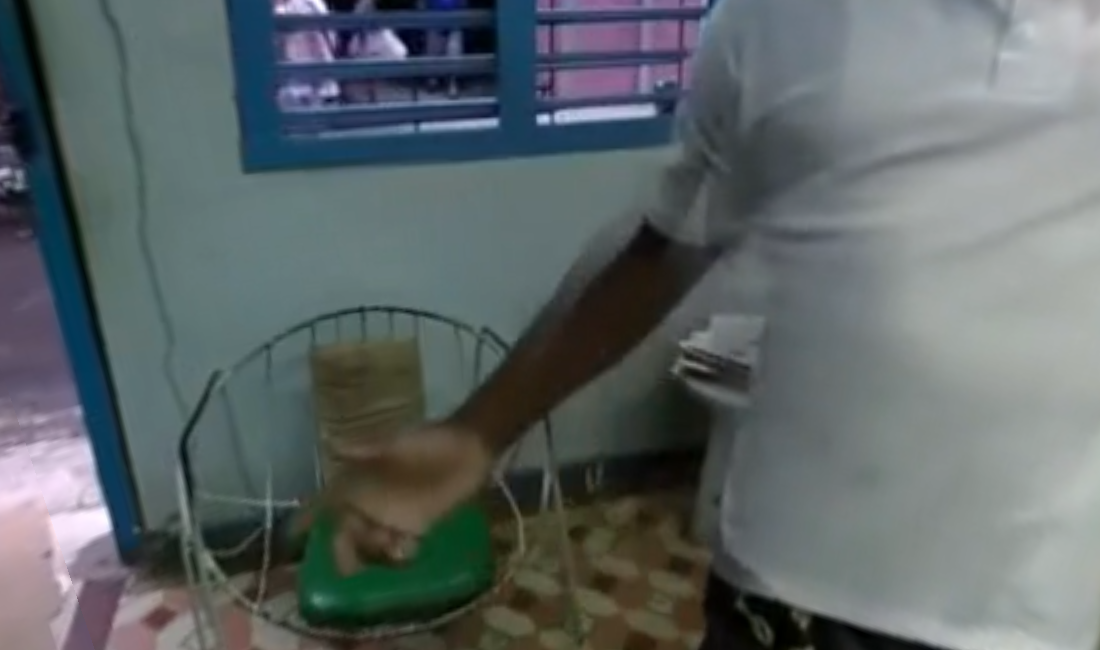}}\hskip.8em
        \subfigure[Artificial poles]{\includegraphics[width=0.225\linewidth]{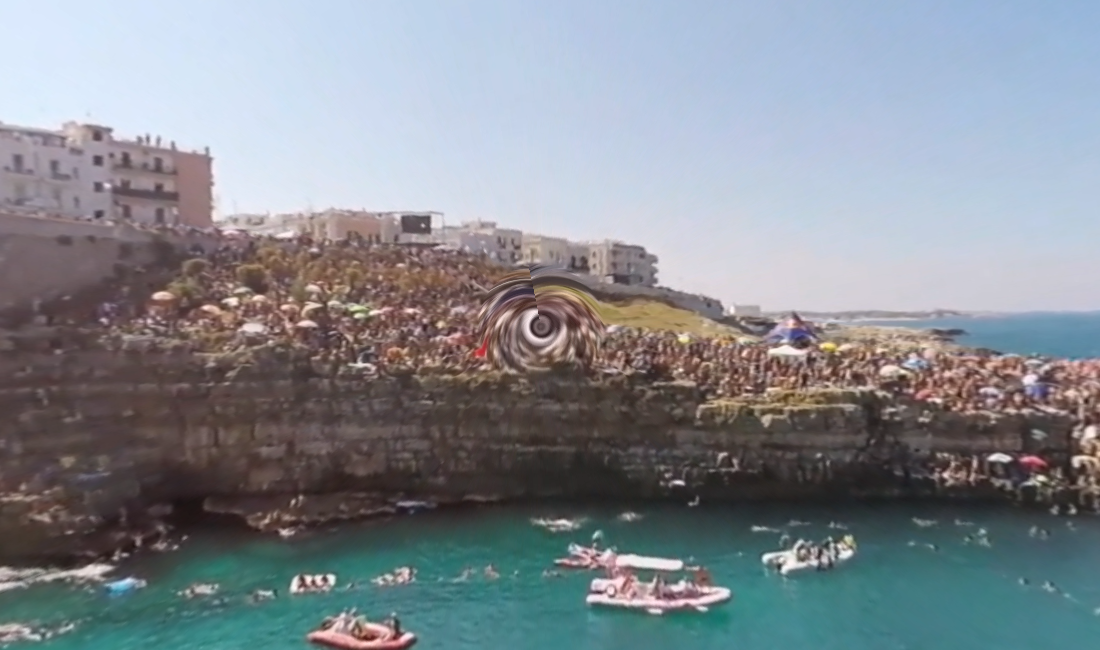}}
        \caption{Visual examples of authentic distortions in VRVQW.}
        \label{fig:VR_distortion} 
\end{figure*}

Different from existing 360{\degree} video databases (refer to Table \ref{tab:summary}), we focus primarily on authentic distortions, which manifest themselves as intricate combinations of multiple visual artifacts that emerge during the creation of 360{\degree} videos ~\cite{Ghadiyaram2016_LIVE}. Fig.~\ref{fig:VR_video_process} shows the entire 360{\degree} video processing pipeline, illustrating that the creation of 360{\degree} videos consists of two steps: optical acquisition using a multi-camera rig and the stitching of multiple planar videos with limited and overlapping FoVs. Visual distortions from the optical acquisition typically result from a combination of factors such as scene complexity, lens imperfection, sensor limitation, and non-professional shooting. These distortions encompass under/over-exposure, out-of-focus and motion blurring, sensor noise, annoying shaky motion, flickering\footnote{Flickering generally refers to unwanted frequent luminance or chrominance changes along the temporal dimension.}, jerkiness\footnote{Jerkiness appears when the temporal resolution is too low to catch up with the speed of moving objects, leading to discontinuous object motion.}, and floating\footnote{Floating denotes the erroneously perceived motion in certain regions relative to their surrounding background, which should remain stationary or move consistently with the background.} \cite{Zeng2014_artifacts}. Stitching distortions are due to the limitation of the stitching algorithm itself and the stitching difficulty introduced by visual distortions from the previous acquisition step (\eg, stitching images of different luminance levels tends to create artificial boundaries, as shown in Fig.~\ref{fig:VR_distortion} (d)). Visually, stitching distortions are abrupt luminance/structure change, objects with missing parts, ghosting, and motion discontinuity localized in space and time. Of particular interest are the artificial converging points visible at the two poles (refer to Fig.~\ref{fig:VR_distortion} (h)). These authentic distortions inevitably affect the whole video processing pipeline and are ultimately perceived by end users.

\textit{2) Viewing Conditions:} We use an HTC Eye Pro to display 360{\degree} videos, which offers an FoV of 110{\degree} and a binocular resolution of $2,880\times1,600$ pixels. Subjects are asked to seat on a swivel chair and wear the HMD to watch the videos. To collect the EM and HM data, we employ the built-in Tobii Pro eye-tracking system with a sampling rate of $2\times$ fps. Video playback is supported by a high-performance server with an AMD Ryzen 9 3950X 16-Core CPU, 128 GB RAM, and an NVIDIA GeForce RTX 2080 Ti GPU. The graphical user interface is customized using the Unity Game Engine. 

\begin{figure}[t]
    \begin{centering}
        \includegraphics[scale = 0.23]{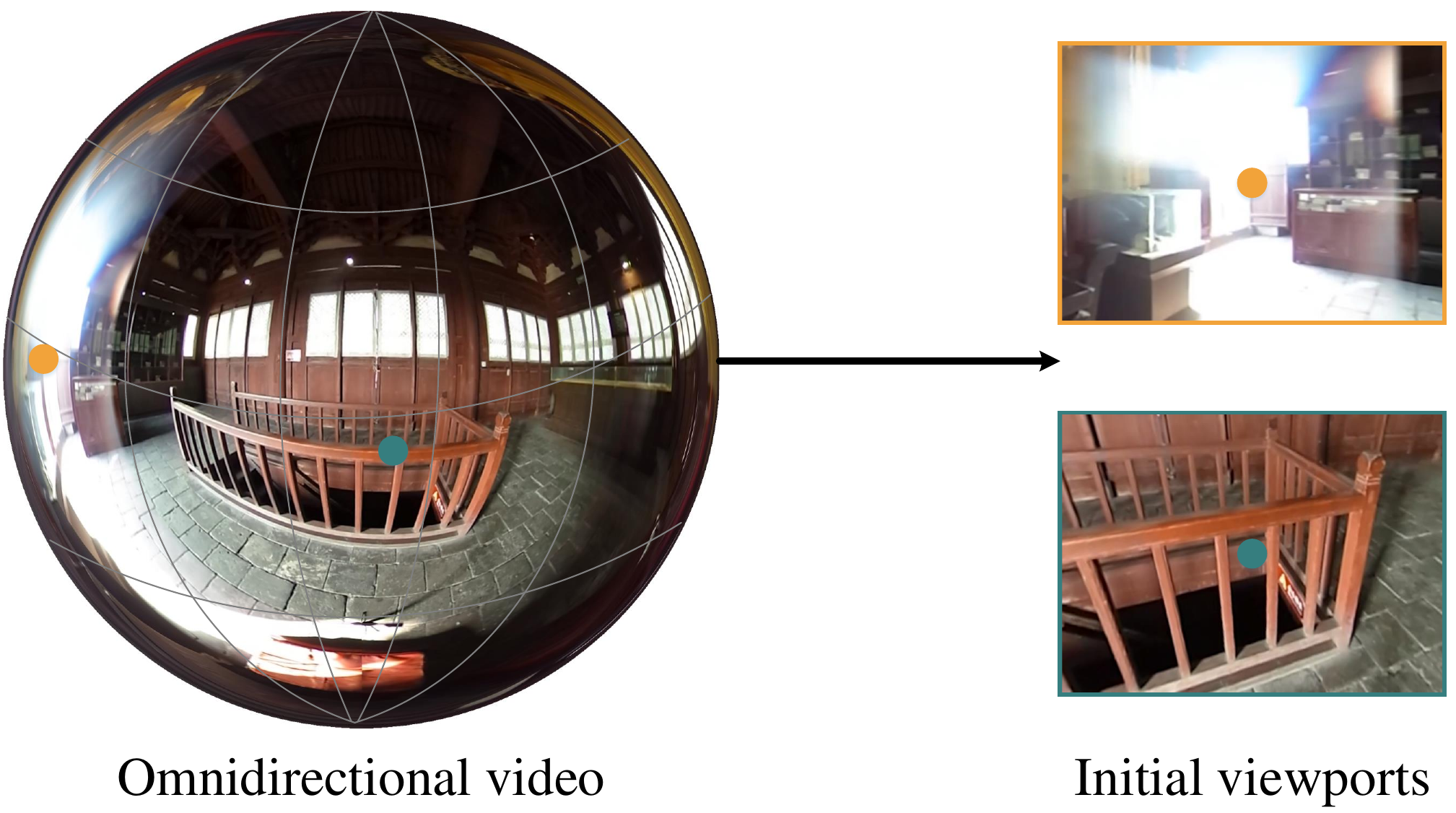}
        \caption{We consider two types of starting points. Starting Point \rom{1} (denoted by the light orange dot) and Starting Point \rom{2} (denoted by the dark green dot) offer poor and good initial viewing experiences, respectively. The video name in VRVQW is ``D\_ConfucianTemple''.}
        \label{fig:starting_point_setting} 
    \end{centering}
\end{figure}

\begin{figure}[t]
    \begin{centering}
        \includegraphics[scale = 0.23]{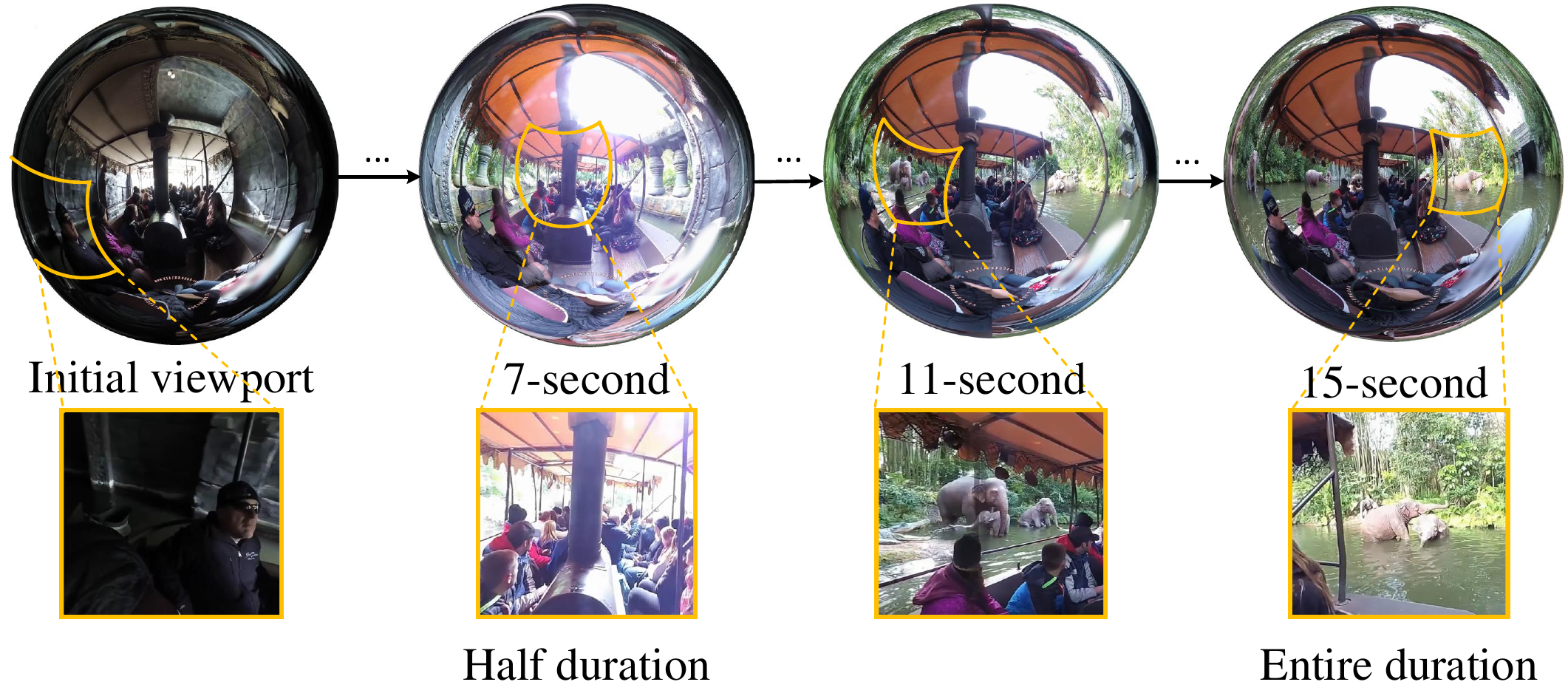}
        \caption{We consider two exploration times, one spanning the entire duration (\ie, 15 seconds) and the other set to the half of the former (\ie, 7 seconds). The initial viewport is from Starting Point \rom{1}. The video name in VRVQW is ``F\_BridegOpening2''.}
        \label{fig:exploration_time_setting} 
    \end{centering}
\end{figure}

VR videos offer an FoV of $360\degree \times 180\degree$, surpassing the natural visual capacity of the human eye. This presents an opportunity to investigate the influence of viewing behaviors on the perceived quality. In real-world scenarios, individuals may watch the same VR video from different starting points, follow different scanpaths, and end the video at various timestamps. As a result, individuals may have varied experiences with the same VR video due to variations in their viewing behaviors. An important consideration in our psychophysical experiment is the variation of two viewing conditions: the starting point and  exploration time.  We intentionally choose \textbf{Starting Point \rom{1}} to give users a \textit{poor} initial viewing experience. This includes viewports that exhibit localized distortions or intensive spatiotemporal information. Another example is the initial viewport from the side when there is strong camera motion guidance. On the contrary, \textbf{Starting Point \rom{2}} is selected to encourage a \textit{good} initial viewing experience and is at least 120{\degree} (in longitude) away from Starting Point \rom{1}. 
An example is shown in Fig.~\ref{fig:starting_point_setting}, where the viewport extracted from Starting Point \rom{1} contains visible over-exposure and color cast distortions, while the viewport extracted from Starting Point \rom{2} is of high quality. Additionally, we establish two exploration times: one spanning the entire duration (\ie, about $15$ seconds) and the other set to half of the former (\ie, $7$ seconds). 
More exploration time allows for the extraction and viewing of more viewports. As shown in Fig.~\ref{fig:exploration_time_setting}, the ``F\_BridegOpening2'' video captures a cruise ship sailing out of a dark cave. When viewing from Starting Point \rom{1} in the first seven seconds, the user may encounter distortions like under-exposure and over-exposure, which can negatively impact her/his viewing experience. In the subsequent eight seconds, the cruise ship has sailed out of the cave, and the viewer may observe artifact-free content, which improves the viewing experience. By combining the starting point and exploration time, we create a total of four different viewing conditions.

\begin{figure}[t]
    \begin{center}
        \includegraphics[scale = 0.3]{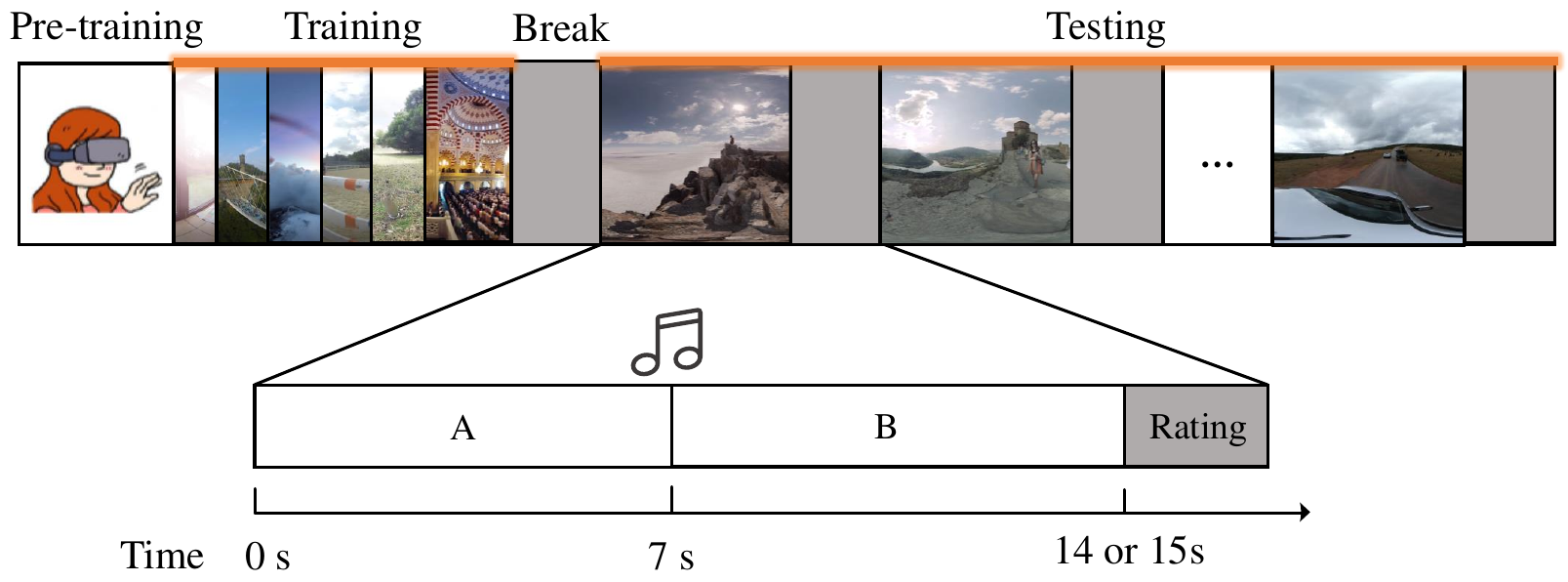}
        \caption{Procedure of our psychophysical experiment. Period A: the first $7$- second viewing. Period B: the second duration of viewing, separated by a voice prompt. After each video sequence is displayed, subjects need to give two scores,  indicating their viewing experience in Period A, and both Periods A and B.}
        \label{fig:experiment_design} 
    \end{center}
\end{figure}

\textit{3) Subjective Methodology:} The psychophysical study employs the single stimulus continuous quality evaluation method described in the ITU-R BT 500.13 recommendation~\cite{series2012methodology}. Subjects are required to rate the perceived quality of a 360{\degree} video on a continuous scale of $[1, 5]$, labeled with five quality levels (``bad'', ``poor'', ``fair'', ``good'', and ``excellent''). To reliably collect MOSs, the experimental procedure consists of three phases: pre-training, training, and testing, as shown in Fig.~\ref{fig:experiment_design}.

In the \textit{pre-training} phase,
basic non-sensitive user information such as age, gender, and the requirement for wearing glasses are recorded. Subjects are familiarized with the experiment's procedure and the rating guideline. We find it relatively time-consuming to teach subjects to use the hand controller for rating. Therefore, one of the authors is responsible for recording the opinion scores, read out by the subjects. 

In the \textit{training} phase,
we select six video sequences that are not included in VRVQW. For the first four videos, subjects are allowed to freely explore the virtual scenes and are informed of the distortions they encounter during exploration. We find that this level of instruction is necessary to familiarize subjects with the distortions that are likely to occur in the testing phase. We try to avoid over-instructing the subjects, such as avoiding providing a reference MOS for each of the four videos. For the remaining two videos, we ask the subjects to give quality scores with no instructions. A discussion on how the subjects arrive at such ratings is held to make sure they understand the evaluation process. No feedback is provided on their scores. Importantly, if the subjects feel any discomfort during this phase, the experiment is interrupted immediately, and they are not invited to participate in the subsequent experiments.

In the \textit{testing} phase, we divide the $502$ videos into eight sessions to reduce fatigue and discomfort that may arise from possible long-time viewing. Additionally, the subjects are allowed to take a break at any time during this phase. Each session contains about $60$ videos, with a $5$-second mid-gray screen in between. We gather human data from $139$ subjects ($75$ females and $64$ males with ages between $17$ and $26$). All participants self-report normal or corrected-to-normal color vision.  The subjects are divided into two groups according to two different sets of starting points. Each subject participates in at least two sessions, and each video is rated by no fewer than $20$ subjects. To collect MOSs for different exploration periods\cite{Duanmu2018ECT}, a voice prompt is played when the subject has viewed half of a 360{\degree} video (about $7$ seconds) to remind her/him of giving a quality score based on the viewing experience so far. After completing the video, s/he is required to give another quality score according to his/her overall viewing experience. It is noteworthy that each video is viewed only once by one subject to ensure that human data is collected without prior knowledge of the scene.

\subsection{Data Processing}

\textit{1) Human Opinion Scores:} After obtaining the raw human scores, we detect and remove outliers using the method in VQEG~\cite{2000VQEG}. Then, we compute the MOS by
\begin{align}
    q_j = \frac{1}{M}\sum_{i=1}^{M}q^{(i)}_j,
\end{align}
where $q^{(i)}_j$ is the opinion score of the $i$-th observer given to the $j$-th video sequence.

\begin{figure}[t]
    \begin{center}
        \includegraphics[scale = 0.65]{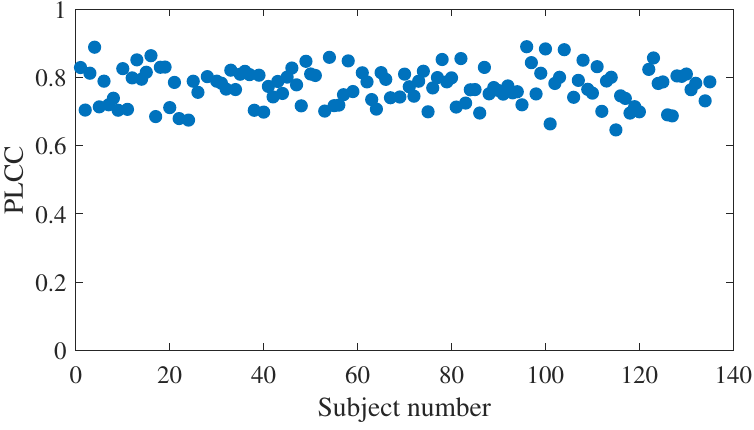}
        \caption{Correlations between ratings from individual subjects and MOSs.}
        \label{fig:reliability_scatter} 
    \end{center}
\end{figure}

To validate the reliability of the collected MOSs in the less controllable VR viewing environment, we calculate the Pearson linear correlation coefficient (PLCC) between the ratings provided by an individual subject and the MOSs.  Fig.~\ref{fig:reliability_scatter} shows the correlation values obtained from $135$ subjects (four subjects are detected as outliers), with a median correlation of $0.784$, which is reasonably high, compared to previous image/video quality databases with authentic distortions~\cite{Hosu2020KonIQ,Ying2021Patch_up}. This result suggests that the adopted subjective rating strategy (with a voice prompt in between) is reliable for collecting MOSs in the VR environment.

\textit{2) Human Viewing Scanpaths:} With the built-in eye-tracking system, we are able to gather both HM and EM data. The HM data for a subject is recorded in the form of a sequence of three Euler angles $[\texttt{pitch},\texttt{yaw}, \texttt{roll}]$. Pitching up/down the head gives a positive/negative $\texttt{pitch}$ value, in the range of $[-90\degree, 90\degree]$; rotating the head to the left/right evokes a positive/negative $\texttt{yaw}$ value, in the range of $[-180\degree, 180\degree]$; tilting the head to the left/right results in a positive/negative $\texttt{roll}$ value. In the context of current VR HMDs, only the $\texttt{pitch}$ and $\texttt{yaw}$ values are relevant, corresponding to the center latitude and longitude coordinates of the extracted viewport. Similarly, the EM data for a subject is directly captured in the form of a sequence of $[\texttt{longitude}, \texttt{latitude}]$, representing the positions at which the eye is looking. The sampling rate for both HM and EM data is $2\times$ fps, with a maximum frequency of $90$ Hz constrained by the HMD. Compared to HM data, EM data tend to be noisier due to the alternating behavioral modes of attention and re-orientation~\cite{Sitzmann2018Saliency_in_VR}. Thus, we define the scanpath of a user as the sequence of $[\texttt{pitch}, \texttt{yaw}]$ derived from the HM data.

\section{Statistical Analysis of VR Data}
\subsection{Understanding Viewing Behaviors in VR}
With the recorded data, we gather insights and investigate a number of questions about user behaviors when watching VR videos in the wild. In this study, we specifically concentrate on analyzing one particular type of viewing behaviors---the scanpath---because it has a significant impact on the perceived quality of a 360{\degree} video by the corresponding user.

\textit{1) Viewing Behavior Metrics:} To compare multiple scanpaths, we adopt two widely used metrics: \textit{temporal correlation}~\cite{anderson2015comparison} and \textit{similarity ring metric}~\cite{Curcio2017_SRM}. We also consider comparing the saliency maps (\ie, the \textit{heatmaps}) as spatial aggregations of the scanpaths for further analysis.

\begin{itemize}

\item \textit{Temporal correlation} (TC) uses the PLCC to calculate the correlations between the longitude and  latitude values of two scanpaths, denoted as $s^{(i)} = [\phi^{(i)},\theta^{(i)}]$ and $s^{(j)}=[\phi^{(j)},\theta^{(j)}]$, followed by simple averaging: 

\begin{align}
\mathrm{TC}\left(s^{(i)}, s^{(j)}\right)=\frac{1}{2} & \left(\operatorname{PLCC}\left(\phi^{(i)}, \phi^{(j)}\right)\right. \\ 
& \left .+\operatorname{PLCC}\left(\theta^{(i)}, \theta^{(j)}\right)\right ), \notag
\end{align}

where $\phi^{(i)}$ and $\theta^{(i)}$ represent the longitudes and latitudes of the $i$-th scanpath, respectively. The mean temporal correlation over $M$ subjects is calculated by
\begin{align}\label{eq:mtc}
    \mathrm{mTC} = \frac{2\sum_{i=1}^{M-1}\sum_{j=i+1}^{M}\mathrm{TC}(s^{(i)},s^{(j)})}{M(M-1)}.
\end{align}

\item \textit{Similarity ring metric} (SRM) measures whether different subjects have been watching the same parts of the video at the same time. While it is less likely for all scanpaths to completely overlap, it is reasonable to determine if they fall within a certain range, \ie, passing through the same \textit{ring}. As suggested by Curcio \etal~\cite{Curcio2017_SRM}, we focus on the longitude of the scanpath and set the radius and the center of the ring as $r = \mathrm{FoV}/2$ and the mode of longitude values from $M$ scanpaths at the same time instance:
\begin{align}
    c_t = \mathrm{mode}\left(\phi^{(1)}_{t}, \phi^{(2)}_{t},\ldots,\phi^{(M)}_{t}\right).
\end{align}
A longitude value out of the ring means that the corresponding subject does not watch the same content with respect to  other subjects at the $t$-th time instance.  The \textit{instantaneous similarity} at the $t$-th time instance for the $i$-th scanpath is then defined as
\begin{align}
    \mathrm{IS}^{(i)}_{t} = 
    \begin{cases}
      1,& \mathrm{if}\ \phi^{(i)}_{t} \in \left[c_{t}-\frac{\mathrm{FoV}}{2}, c_{t}+\frac{\mathrm{FoV}}{2}\right],\\
      0,& \mathrm{otherwise},  
    \end{cases}
\end{align}
based on which we compute the SRM by averaging across all scanpaths and over all time instances:
\begin{align}\label{eq:srm}
    \mathrm{SRM} = \frac{100}{MT}\sum_{t=1}^T\sum_{i=1}^{M}\mathrm{IS}^{(i)}_{t}.
\end{align}
SRM is scaled to lie within $[0,100]$, and a larger value indicates  higher consistency.

\item \textit{Heatmap} reflects the salient areas to which users pay attention and can be considered as a spatial aggregation of the scanpaths. To generate dynamic heatmaps for omnidirectional videos, we apply the density-based spatial clustering (DBSCAN) algorithm~\cite{Ester1996_DBSCAN} on the scanpaths of all subjects. Fixations are defined as the cluster centroids that span at least $200$ ms~\cite{Chao2020_Audio-Visual}, during which the gaze direction remains relatively unchanged.  Noisy fixation points will be automatically filtered out. For every second of the video sequence, we compute a fixation map by DBSCAN. The saliency of each location in the fixation map is determined by the total spherical (\ie, great-circle) distances from the location to all fixation points~\cite{Xu2019HeadPrediction}, normalized by the computed maximum distance. To compare the similarity of two heatmaps, we follow Sitzmann \etal~\cite{Sitzmann2018Saliency_in_VR}, and use PLCC as the quantitative measure.

\end{itemize}

\begin{table}[t]
\caption{Viewing behavior consistency in terms of $\mathrm{m}$TC and SRM (and the standard error) under different viewing conditions\cite{vrfang2023}}
\label{table:scanpath_plcc_srm}
\centering
\renewcommand\arraystretch{1.25}
\begin{tabular}{cl|c|c}
\whline
\multicolumn{1}{l}{}                       &           & \multicolumn{1}{l|}{Starting Point \rom{1}} & Starting Point \rom{2} \\ \hline
\multicolumn{1}{c|}{\multirow{2}{*}{mTC}} & 7-second  & 0.394 ($\pm$0.046)                            & 0.395 ($\pm$0.041)       \\
\multicolumn{1}{c|}{}                      & 15-second & 0.289 ($\pm$0.038)                            & 0.286 ($\pm$0.034)       \\ \hline
\multicolumn{1}{c|}{\multirow{2}{*}{SRM}}  & 7-second  & 72.997 ($\pm$7.221)                            & 74.702 ($\pm$7.670)      \\
\multicolumn{1}{c|}{}                      & 15-second & 63.721 ($\pm$5.371)                           & 65.128 ($\pm$5.603)     \\ 
\whline
\end{tabular}
\end{table}

\begin{figure}[t]
    \begin{center}
        \includegraphics[scale = 0.5]{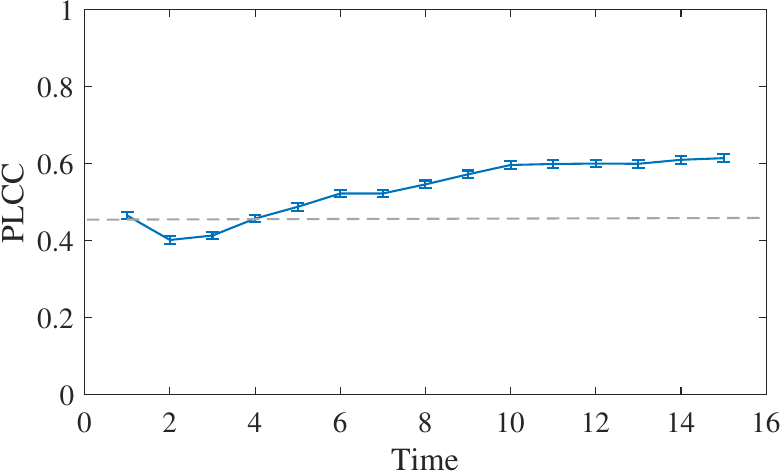}
        \caption{Viewing behavior consistency in terms of PLCC between heatmaps from Starting Point I and Starting Point II, averaged over all 360{\degree} videos. The dashed line represents the initial PLCC\cite{vrfang2023}.}
        \label{fig:CC_scater} 
    \end{center}
\end{figure}

\textit{2) Does the Viewing Condition Affect Viewing Behaviors?} To assess whether viewing behaviors are affected by the viewing conditions, we calculate mTC and SRM  under different viewing conditions, as listed in Table~\ref{table:scanpath_plcc_srm}. We also employ the analysis of variance (ANOVA) to see whether such differences in viewing behavior consistency as measured by mTC and SRM are statistically significant, as listed in Table~\ref{tab:anova_on_plcc_srm}. 
From the tables, we find that human viewing behavior consistency, as measured \textit{globally} by mTC,  is relatively low across all viewing conditions. However, when measured more \textit{locally} by SRM, the consistency improves, and the differences in consistency for different starting points and exploration times are statistically significant, as evidenced by $p$-values close to zero.  

Furthermore, we compute the mean PLCC values between the heatmaps from Starting Point \rom{1} and Starting Point \rom{2}, averaged over different videos, as shown in Fig.~\ref{fig:CC_scater}.  We find that, within the first four seconds, the PLCC is lower than the initial, indicating divergent heatmaps, which can be explained by  viewing conditions.  The PLCC increases over time, but does not reach a sufficiently high level to eliminate the impact of viewing conditions.

\begin{table}[t]
\caption{$p$-values in the ANOVA test for $\mathrm{m}$TC and SRM. A \textit{p}-value below the threshold of $0.05$ represents that the corresponding factor has a significant impact on viewing behavior consistency, as measured by mTC and SRM}
\label{tab:anova_on_plcc_srm}
\renewcommand\arraystretch{1.25}
\centering
\begin{tabular}{l|c|c}
\whline
Factor            & mTC    & SRM   \\ \hline
Starting point     & 0.615 & $\approx$ 0     \\
Exploration time   & $\approx$ 0     & $\approx$ 0     \\
\whline
\end{tabular}
\end{table}

\begin{table}[t]
\caption{Perceived quality analysis in terms of MOS (and the standard error) under different viewing conditions\cite{vrfang2023}}
\label{table:MOSs_mean_error}
\renewcommand\arraystretch{1.25}
\centering
\renewcommand\arraystretch{1.25}
\renewcommand\tabcolsep{8pt}
\begin{tabular}{l|l|l}
\whline
          & Starting Point \rom{1} & Starting Point \rom{2} \\ \hline
7-second  & 2.551 ($\pm$ 0.034)           & 3.059 ($\pm$ 0.034)            \\ \hline
15-second & 3.119 ($\pm$ 0.036)          & 2.562 ($\pm$ 0.038)            \\ 
\whline
\end{tabular}
\end{table}

\subsection{Understanding Perceived Quality in VR}
\label{sec:sec5}
Understanding how individuals perceive visual distortions in VR environments poses challenges due to the differences in viewing conditions between planar and immersive 360{\degree} videos. In this subsection, we analyze the effects of VR viewing conditions and video attributes on the perceived quality of 360{\degree}  videos.

\begin{table}[t]
\caption{Perceived quality analysis in terms of MOS (and the associated standard error) under different video attributes}
\label{table:attributes_analysis}
\centering
\renewcommand\arraystretch{1.25}
\begin{tabular}{l|c}
\whline
\multicolumn{1}{l|}{Video attribute} & Average MOS \\ \hline
Low-resolution                        & 2.176 ($\pm$ 0.024) \\
High-resolution                       & 3.081 ($\pm$ 0.021)\\ 
No camera motion                      & 2.856 ($\pm$ 0.021)\\ 
Camera motion                         & 2.779 ($\pm$ 0.037)\\ 
\whline
\end{tabular}
\end{table}

\textit{1) Does the Viewing Condition Affect Perceived Quality?} Previous studies~\cite{Singla2017OVD,Curcio2017OVD,Tran2017OVD,IVQAD2017,Zhang2018_subjective,ZhangY2018OVD,Lopes2018subjective,li2018_subjective,Meng2021OVD} assume that viewing conditions have a negligible impact on the perceived quality of 360{\degree} videos. This assumption holds true when considering \textit{synthetic} artifacts (\eg, video compression) with \textit{global} distortion appearances. In such cases, regardless of head orientation or any time instance, the extracted viewport is highly likely to contain the same degree of artifacts. However, this is not the case when it comes to VR videos in the wild, where we are dealing with \textit{authentic} distortions, \textit{localized} in space and time. Whether and when to encounter such spatiotemporal local distortions may have a different influence on the perceived quality. To test the hypothesis, we average the MOSs in VRVQW for different viewing conditions in Table~\ref{table:MOSs_mean_error}.  Several useful findings are worth mentioning. First,  compared to a $15$-second exploration, a $7$-second exploration allows for fewer viewports of the scene to be observed, highlighting the importance of the starting point to the perceived video quality. Second, when longer exploration time is allowed,  viewports closer to the end of the video exert a greater influence on the overall viewing experience due to the \textit{recency effect}~\cite{Hands2001recency_effect}.
This explains why, for the $7$-second exploration, subjects tend to give low-quality scores when viewing from Starting Point I, where distortions appear in the initial viewports (see Sec. \ref{subsec:dg} for the definitions of the two types of starting points). Given more time, the subjects would consciously re-orient their heads to avoid viewing distorted viewports and seek those with better quality, leading to an improved viewing experience. On the contrary, from Starting Point II, where the distortions may not be viewed initially, the subjects are less likely to observe distortions within the limited exploration time, thus explaining the higher average MOS of the $7$-second exploration.

\begin{table}[t]
\caption{Multi-factorial ANOVA test results for the effects of the starting point,  exploration time,  spatial resolution, and  camera motion on the \textit{perceived quality}. $SS$:  sum of squares. $d. f.$: degrees of freedom. $MS$: mean square. $F$: $F$-value. $p$: $p$-value for the null hypothesis. We omit three- and four-factorial analysis results, which are statistically insignificant}
\label{table:anova_on_qa}
\renewcommand\arraystretch{1.25}
\centering
\begin{tabular}{l|l|l|l|l|l}
\whline
\textit{Source of variation }   & $SS$  & $d.f.$  & $MS$  & $F$  & $p$    \\ \hline
Starting point              & 1.600    & 1    & 1.604  & 3.560   & 0.059 \\ \hline
Exploration time            & 0    & 1    & 0.004  & 0.010   & 0.927 \\ \hline
Spatial resolution          & 207.250   & 1    & 207.250 & 459.79 & $\approx$ 0 \\ \hline
Camera motion               & 0.200    & 1    & 0.201  & 0.450   & 0.505 \\ \hline
\begin{tabular}[c]{@{}l@{}}Starting point \\ Exploration time  \end{tabular} & 71.360   & 1    & 71.362 & 158.32  & $\approx$ 0 \\ \hline
\begin{tabular}[c]{@{}l@{}}Starting point \\ Spatial resolution  \end{tabular} & 0.300  & 1    & 0.295 & 0.650  & 0.419 \\ \hline
\begin{tabular}[c]{@{}l@{}}Starting point \\ Camera motion  \end{tabular} & 0.720  & 1    & 0.716 & 1.590  & 0.208 \\ \hline
\begin{tabular}[c]{@{}l@{}}Exploration time \\ Spatial resolution  \end{tabular} & 0.110 & 1    & 0.105 & 0.230  & 0.629\\ \hline
\begin{tabular}[c]{@{}l@{}}Exploration time \\ Camera motion  \end{tabular} & 0.890 & 1    & 0.888  & 1.970  & 0.161\\ \hline
\begin{tabular}[c]{@{}l@{}}Spatial resolution \\ Camera motion \end{tabular}  & 3.750    & 1    & 3.746  & 8.310   & 0.004 \\ \hline
Error   & 897.900 & 1992 & 0.451  &         &       \\ \hline
Total   & 1386.140 & 2007 &        &         &       \\ 
\whline
\end{tabular}
\end{table}

\textit{2) Does the Video Attribute Affect Perceived Quality?} We consider two video attributes: camera motion and spatial resolution. 
The camera motion is typically generated by body-mounted cameras. 
We carefully select $32$ videos featuring complex camera motion (\eg, camera mounted on the roller coaster or held by a surfer). 
From Table~\ref{table:attributes_analysis}, we find, as expected, that videos with camera motion generally receive lower MOSs regardless of  the viewing conditions. This validates that using camera motion as strong visual guidance often results in annoying shaky motion, impairing the user viewing experience. We also show in Table~\ref{table:attributes_analysis} that high-resolution videos receive a higher average MOS than low-resolution ones (for all viewing conditions), even after being downsampled in the HMD. It remains to be determined whether such downsampling has a positive impact on the perceived quality by ``concealing'' certain types of distortions (\eg, compression artifacts and high-frequency noise).

\begin{figure*}[t]
    \begin{center}
       \includegraphics[scale = 0.5]{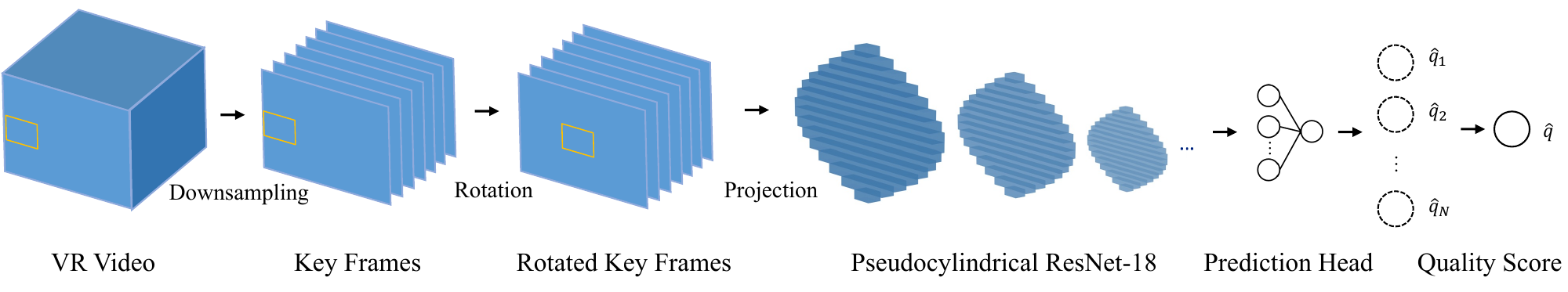}
        \caption{System diagram of the proposed VR VQA model. Our model consists of four components: 1) a preprocessor for spatiotemporal downsampling and rotation, accounting for the two viewing conditions, 2) a projector for ERP to pseudocylindrical representation conversion, mitigating geometric distortions by ERP, 3) a feature extractor implemented by   
        a pseudocylindrical CNN for spatial feature analysis, and 4) a quality regressor to compute frame-level scores, followed by temporal pooling.}
        \label{fig:TSVQ}
    \end{center}
\end{figure*}

\textit{3) Significant Impact Analysis:} To test the statistical significance of the four factors --- the starting point, exploration time,  spatial resolution, and  camera motion --- on the perceived quality, we apply the multi-factorial ANOVA to the MOSs among factors. 
Multi-factorial ANOVA is a statistical analysis technique used to examine the effects of multiple independent variables on a dependent variable.
It starts by specifying a null hypothesis that the analyzed factors have no effect on the outcome. If the calculated $p$-value is less than $0.05$, we reject the null hypothesis, meaning that the analyzed factors significantly influence the perceived quality.
The results are listed in Table~\ref{table:anova_on_qa}, where we confirm that the spatial resolution is a significant \textit{individual} factor. The effect of the camera motion alone is not statistically significant, partially due to the limited inclusion of such videos to avoid visual discomfort. The perceived quality of user-generated VR videos is determined by the combination of two viewing conditions (\ie, the starting point and exploration time).

\section{Objective Quality Assessment of VR Videos in the Wild}
In this section, we describe a blind  VQA model  that 1) accounts for the two viewing conditions (\ie, the starting point and  exploration time),  2) respects the spherical nature of VR videos, 3) delivers accurate quality prediction performance, and 4) is computationally efficient.

\subsection{Overview on Model Challenges}
To the best of our knowledge, there is currently no objective VQA model specifically designed for VR videos in the wild. Such a proper quality model  must address several key issues.

\begin{itemize}

\item \textit{Spatial Resolution}.  VR videos encompass the entire 360{\degree}$\times$180{\degree} FoV with high fidelity, resulting in an exceedingly high spatial resolution. To accelerate computational prediction, it is necessary to consider spatial
downsampling as one of the preprocessing steps. But, how should we
determine the downsampling factor?

\item \textit{Frame Rate}. Similar to the spatial resolution, a high frame rate is essential for VR videos to enhance the sense of presence and to minimize motion sickness. For computational reasons, should we also perform temporal downsampling to identify and operate solely on a set of  key frames?

\item \textit{Viewing Condition}. Na\"{i}ve application of planar VQA models to VR videos under different viewing conditions is problematic because the same video is accompanied by multiple MOSs, each associated with one viewing condition. How to naturally model viewing conditions is the key to the success of VR VQA in the wild, but is little investigated.

\item \textit{Spherical Geometry}.  VR videos are typically stored in the ERP format, which introduces severe geometric distortions, especially in the vicinity of poles. Thus,
 directly transferring planar I/VQA models may achieve suboptimal performance. Among many sphere-to-plane and plane-to-sphere projections, which one should be adopted in VR VQA?

\end{itemize}

\begin{figure}[t]
    \begin{centering}
        \includegraphics[scale = 0.7]{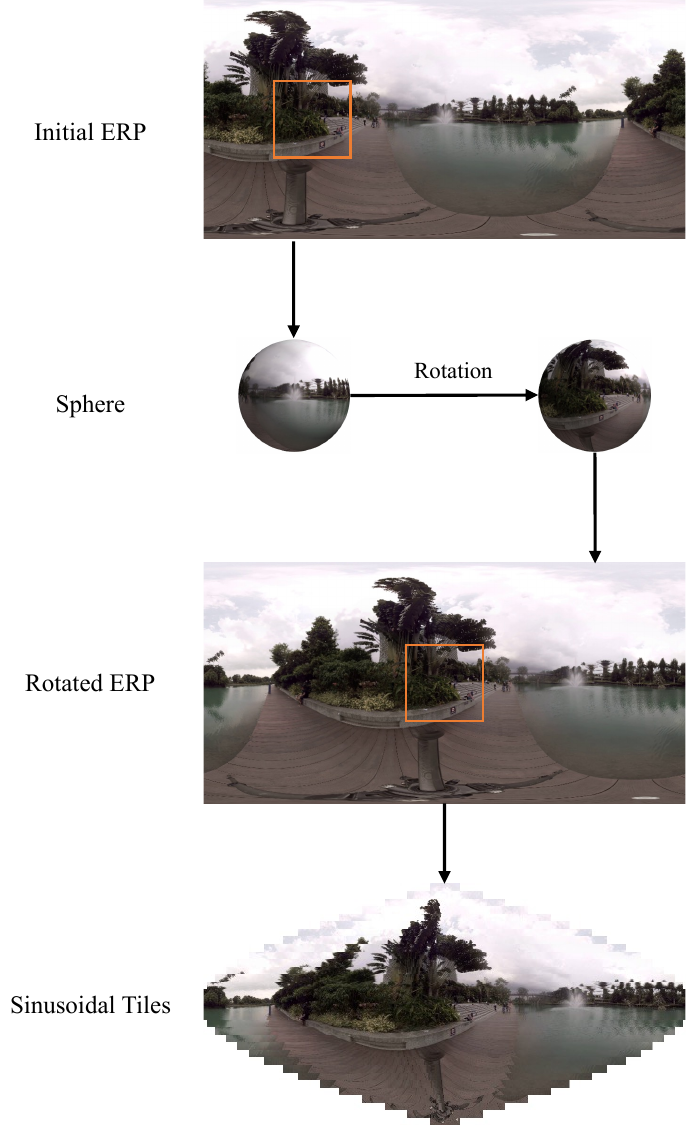}
        \caption{Illustration of rotation and projection. We start by projecting an ERP frame onto a unit sphere,  rotating it to center the initial viewport specified by the starting point, and  projecting it back to the ERP format. We then transform the rotated frame into a tiled sinusoidal representation as a special case of the pseudocylindrical representation. The video name in VRVQW is ``A\_Bay''.}
        \label{fig:rotate} 
    \end{centering}
\end{figure}

\subsection{Proposed Method}
To confront the above-mentioned model design challenges, we describe a VR VQA model encompassing four basic building blocks: 1) a preprocessor, 2) a projector, 3) a feature extractor, and 4) a quality regressor. The system diagram of our method is illustrated in Fig.~\ref{fig:TSVQ}.

\begin{itemize}

\item \textit{Preprocessor}. As mentioned earlier, VR videos often come with high resolutions and frame rates, which desire spatiotemporal downsampling. Our computational analysis reveals that the spatial distortions  occurring in VR videos are fairly visually stable under spatial downsampling, and do not affect subsequent feature extraction. Thus, bilinear interpolation is adopted to downsample the ERP frames to a spatial resolution of $1,024 \times 512$. For temporal downsampling, it can be leveraged to model the exploration time viewing condition. Specifically, we may set the temporal downsampling rate to be linearly proportional to the exploration time. In our settings, we choose to downsample the $7$-second and $15$-second videos in VRVQW to $1$ and $0.5$ fps, both leading to $7$ key frames per video. To accommodate the starting point as the second viewing condition, we rotate the starting point to $(\phi,\theta) = (0,0)$ with the minimum geometric distortion in  ERP, as illustrated in Fig.~\ref{fig:rotate}.

\item \textit{Projector}. According to Gauss's Theorema Egregium, any projection from a sphere to a plane inherently introduces some distortions. Thus, it is preferable to perform computation directly on the sphere. However, spherical operators (\eg, spherical convolution) are often computationally intensive, making them less suitable for time-sensitive video applications like VQA. Recently, Li \etal~\cite{li2021pseudocylindrical} proposed the pseudocylindrical convolution as an effective approximation to spherical convolution, while enjoying efficient implementation by standard convolution. The pseudocylindrical convolution needs to operate on the so-called pseudocylindrical representation~\cite{li2021pseudocylindrical}. Here we adopt a special case of the pseudocylindrical representation---sinusoidal tiles~\cite{yu2015content}---as the projection format. Fig.~\ref{fig:rotate} illustrates the geometric distortions in the ERP, which are more pronounced with the increasing latitude. By dividing and scaling the ERP into sinusoidal tiles, the geometric distortions are largely reduced, while keeping a uniform sampling rate across
different latitudes. Moreover, pseudocylindrical convolution (\ie, standard convolution with pseudocylindrical padding\footnote{Given the current tile, we pad the latitudinal side with adjacent tiles
resized to the same width, and pad the longitudinal side circularly to respect the spherical structure.}) can be directly applied to the sinusoidal tiles.  Assuming a VR video frame $x \in \mathbb{R}^{H \times W}$, where $H$ and $W$ represent the height and  width, respectively,  we divide and scale $x$ into $S$ tiles (\ie, nonoverlapping rectangles, see Fig.~\ref{fig:rotate}), $\{x_i\}_{i=1}^S$, where $x_i\in\mathbb{R}^{H_i\times W_i}$, $H_i= H/S$, $W_i=\cos(\theta_i)W$, and $\theta_i$ is the (mean) latitude of the $i$-th tile. 

\begin{figure*}[t]
    \begin{center}
       \includegraphics[scale = 0.55]{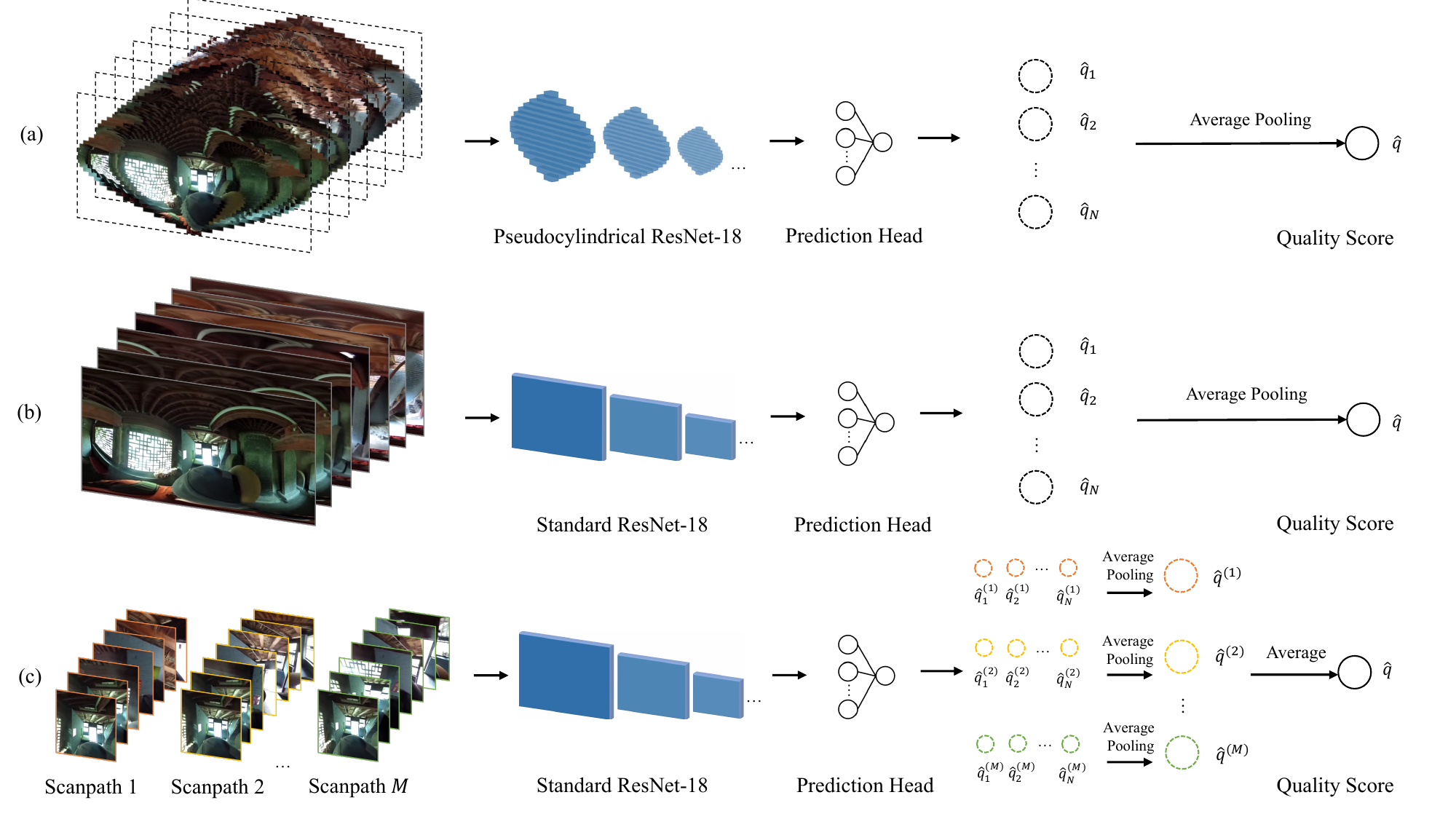}
        \caption{Comparison of (a) the proposed method with its two variants: (b) ERP-VQA and (c) Scanpath-VQA. $N$ represents the number of key frames after preprocessing. $M$ represents the number of scanpaths used in Scanpath-VQA.}
        \label{fig:vqa_models}
    \end{center}
\end{figure*}

\item \textit{Feature Extractor}. We choose a variant of ResNet-18~\cite{He2016ResNet} pre-trained on ImageNet~\cite{deng2009imagenet} as the spatial feature extractor. We make minimal modifications to ResNet-18 by striping the classification head  and replacing all standard convolution with pseudocylindrical convolution. To define the pseudocylindrical convolution, we need to first specify the neighboring grid of $(u, v)\in\{1,\cdots, H\}\times\{1,\cdots, W\}$:
\begin{align}
\mathcal{N}=\{(i, j) \mid i, j \in\{-K, \ldots, K\}\},
\end{align}
from which we  the neighbors through the sphere-to-plane projection~\cite{li2021pseudocylindrical}:
\begin{align}
u_i & =u+i, \\ \label{eq:q}
v_j & =\frac{W_{u_i}}{W_u}(v+0.5)-0.5+\frac{\cos \theta_u}{\cos \theta_{u_i}} \frac{W_{u_i}}{W_u} j \\  \notag 
& =\frac{W_{u_i}}{W_u}\left(v+\frac{\cos \theta_u}{\cos \theta_{u_i}} j+0.5\right)-0.5 .
\end{align}
As $u$ and $u_i$ are close provided that $H$ is large, we may assume $\cos \left(\theta_u\right) \approx \cos \left(\theta_{u_i}\right)$ and simplify Eq.~\eqref{eq:q} to
\begin{align}\label{eq:q2}
v_j \approx \frac{W_{u_i}}{W_u}(v+j+0.5)-0.5,
\end{align}
from which we find that adjacent samples in a tile are just neighbors of one another. Moreover,
searching for neighbors in an adjacent tile amounts to scaling it to the width of the current tile. This is referred to as pseudocylindrical padding~\cite{li2021pseudocylindrical}. Pseudocylindrical convolution over sinusoidal tiles (or the more general pseudocylindrical representation) can be efficiently implemented by standard convolution with pseudocylindrical padding. The adoption of pseudocylindrical convolution brings a significant advantage to the VR VQA model design: advanced model architectures as well as pre-trained weights for planar images and videos can be directly inherited with no modifications.

\end{itemize}

\begin{itemize}

\item \textit{Quality Regressor}. 
After extracting the feature representation for each key frame, we adopt a linear prediction head to regress frame-level quality scores. The last step is to aggregate frame-level quality scores into an overall quality estimate. Among various temporal pooling strategies~\cite{Tu2020Temporal_pooling}, we choose the simple average pooling.

\end{itemize}

\section{Evaluating VQA Models on VRVQW}
In this section, we compare the proposed VR VQA model with existing blind I/VQA models as well as two of its variants, as the performance ``lower bound'' and ``upper bound'', respectively (see Fig.~\ref{fig:vqa_models}).

\begin{table*}[t]
\caption{Performance comparison of different I/VQA methods on the proposed VRVQW under different viewing conditions. The top-$2$ results are highlighted in bold}
\label{table: qa_performance_sp}
\renewcommand{\arraystretch}{1.25}
\centering
\resizebox{1\textwidth}{!}{
\begin{tabular}{l|ccccccc|ccccccc}

\whline
\multirow{3}{*}{Model} & \multicolumn{7}{c|}{SRCC$~\uparrow$} & \multicolumn{7}{c}{PLCC$~\uparrow$}  
\\ \cline{2-15} 
                       & \multicolumn{2}{l}{Starting Point \rom{1}} & \multicolumn{2}{l}{Starting Point \rom{2}} & \multicolumn{2}{l}{Starting Point \rom{1}\&\rom{2}} & \multicolumn{1}{c|}{\multirow{2}{*}{Overall}} 
                       & \multicolumn{2}{l}{Starting Point \rom{1}} & \multicolumn{2}{l}{Starting Point \rom{2}} & \multicolumn{2}{l}{Starting Point \rom{1}\&\rom{2}} & \multicolumn{1}{c}{\multirow{2}{*}{Overall}} \\
                       & 7s                & 15s              & 7s                & 15s     & 7s                & 15s    & \multicolumn{1}{c|}{}     & 7s                & 15s              & 7s                & 15s     & 7s                & 15s & \multicolumn{1}{c}{} \\ \hline
NIQE & 0.350 & 0.455 & 0.461 & 0.429 & 0.387 & 0.428 & 0.407 & 0.322 & 0.417 & 0.409 & 0.388 & 0.348 & 0.393 & 0.371 \\
VSFA  & 0.694 & 0.791 & 0.824 & 0.813 & 0.684 & \textbf{0.788} & 0.750 & 0.720 & 0.809 & 0.825 & 0.814 & 0.694 & \textbf{0.792} & 0.746\\
Li22  & \textbf{0.764} & 0.830 & \textbf{0.853} & 0.859 & 0.642 & 0.505 & 0.438  & \textbf{0.780} & 0.842 & \textbf{0.858} & \textbf{0.881} & 0.664 & 0.520 & 0.508 \\
MC360IQA & 0.738 & 0.771 & 0.762 & 0.798 & 0.652 & 0.675 & 0.672& 0.734 & 0.778 & 0.760 & 0.799 & 0.653 & 0.676 & 0.669 \\ 
\hline
ERP-VQA & 0.712 & 0.849 & 0.847 & \textbf{0.865} & 0.723 & 0.782 & 0.751 & 0.738 & 0.857 & 0.846 & 0.874 & 0.730 & 0.791 & 0.753 \\
Scanpath-VQA & \textbf{0.772} & \textbf{0.865} & 0.813 & 0.850 & \textbf{0.785} & \textbf{0.787} & \textbf{0.781} & \textbf{0.782} & \textbf{0.874} & 0.836 & 0.858 & \textbf{0.785} & 0.789 & \textbf{0.783} \\
\hline
Proposed & 0.730 & \textbf{0.862} & \textbf{0.859} & \textbf{0.872} & \textbf{0.737} & 0.784 & \textbf{0.757} & 0.755 & \textbf{0.868} & \textbf{0.860} & \textbf{0.881} & \textbf{0.738} & \textbf{0.794} & \textbf{0.761} \\
\whline
\end{tabular}
}
\end{table*}

\subsection{Model Selection}

We select and adapt six blind I/VQA models, as there is currently no blind VR VQA model for direct comparison.
\begin{enumerate}

\item NIQE~\cite{Mittal2013NIQE}, the Natural Image Quality Evaluator, is a completely blind IQA model that does not rely on MOSs for training. NIQE measures the deviation of the test image from statistical regularities of natural undistorted images. 

\item VSFA~\cite{Li2019VSFA}, the Video Semantic Feature Aggregation, is a blind VQA model with content-aware feature extraction and temporal memory modeling. 

\item Li22~\cite{li2022blindly}, a two-stream CNN-based VQA model, leverages domain knowledge from spatial appearance and temporal motion through transfer learning. Li22 employs a pre-trained CNN for IQA to extract spatial features and a pre-trained CNN for action recognition to extract motion features. These features are  fed into a quality regressor, guided by the Spearman rank-order correlation coefficient (SRCC) and PLCC as the hybrid loss. Notably, Li22 sets the performance record for planar VQA.

\item MC360IQA~\cite{Sun2020MC360IQA}, the Multi-Channel CNN for blind 360{\degree} IQA, uses six viewports covering the panoramic scene as input. Six parallel hyper-ResNet-34 networks~\cite{He2016ResNet} are used to extract features, which are then concatenated and fed into a quality regressor.

\item ERP-VQA, a special case of the proposed method, directly takes ERP frames as input, processed by ResNet-18 with standard convolution. The preprocessor and the quality regressor remain the same. Conceptually, ERQ-VQA gives the performance ``lower bound'' for the proposed method.

\item Scanpath-VQA, another variant of the proposed method, extracts a set of $224\times 224$ planar videos (corresponding to an FoV of $90{\degree}\times 90{\degree}$) by sampling, along users' scanpaths, a number of viewports from the VR video. Since viewing behaviors under different viewing conditions are explicitly modeled, the preprocessor of Scanpath-VQA performs spatiotemporal downsampling only. The remaining processing for each extracted planar video is the same as ERP-VQA. The overall quality score is computed by averaging 2D video quality estimates:
     \begin{align}\label{eq:qe}
         \hat{q} = \frac{\sum_{i=1}^M \hat{q}^{(i)}}{M},
     \end{align}
 where $M$ is the number of scanpaths adopted for planar video extraction. Eq.~\eqref{eq:qe} can be seen as a quality ensemble, which has the potential to boost the prediction performance. However, the computational complexity of Scanpath-VQA increases roughly by a factor of $M$, making it less practical when $M$ is large. More importantly, Scanpath-VQA relies on the ground-truth scanpaths for quality computation, and  thus  provides the performance ``upper bound'' for our method. 
\end{enumerate}

\subsection{Implementation Details}
The implementations of all competing models are obtained from the original authors. 
For NIQE that does not require training, we compute per-ERP-frame quality score without spatiotemporal downsampling nor rotation, followed by simple average pooling.
For VSFA and Li22, we re-train two separate models for two starting points in the ERP domain without incorporating the preprocessor, and combine the results of the two models for evaluation. 
For MC360IQA as an end-to-end fine-tuned blind IQA model, we adopt the same preprocessor as Scanpath-VQA, and re-train it on VRVQW by assigning the video-level quality score to each key frame. The temporally averaged score is utilized for testing.
For ERP-VQA, Scanpath-VQA and the proposed method, all learnable parameters are optimized for the PLCC loss using the Adam optimizer with a minibatch size of $8$ and a learning rate of $5 \times 10^{-5}$  for $30$ epochs. The number of scanpaths, $M$, in Scanpath-VQA is set to $20$, provided in VRVQW.

We randomly split VRVQW  into three non-overlapping sets: $60\%$ for training, $20\%$ for validation, and $20\%$ for testing. To avoid any bias caused by random partitioning, we repeat this process $10$ times, and report the median results.

\subsection{Performance Comparison}

We summarize the SRCC and PLCC results in Table~\ref{table: qa_performance_sp}, from which we have several interesting observations. 
First, the model performance for Starting Point \rom{2} is generally better than that for Starting Point \rom{1}. 
Such inaccuracy is more pronounced when the viewing time is shorter.
Second, as expected, Scanpath-VQA sets the performance upper bound for all methods. Of particular interest is its comparison to MC360IQA, which essentially uses a set of synthetic still scanpaths for planar video extraction. The improved performance of Scanpath-VQA emphasizes  the importance of incorporating ground-truth (or accurately predicted) users scanpaths into VR VQA. 
Third, our method ranks second, and consistently outperforms ERP-VQA, highlights the role of sinusoidal tiles and pseudocylindrical convolution to mitigate geometric distortions in the ERP format and to construct effective VR VQA models.
Fourth, the reasonable performance by VSFA is  due primarily to the training of two separate models for different viewing conditions. When the training pipeline in the original paper is applied, the performance drops significantly. Nevertheless, its inferior performance relative to our method shows the necessity of  end-to-end training/fine-tuning. 
Last but not least, it should be stressed that even Scanpath-VQA, allowing the use of ground-truth scanpaths, cannot consistently provide the best performance across all viewing conditions, indicating ample opportunities to further advance blind VR VQA.

\begin{table}[t]
\caption{Performance comparison of different I/VQA methods on BIT360 and VQA-ODV.``-'' means that the corresponding model is not applicable}
\label{table:qa_performance_bit360_vqaodv}
\renewcommand\arraystretch{1.25}
\centering
\begin{tabular}{l|cc|cc}
\whline
 \multirow{2}{*}{Method}  & \multicolumn{2}{c|}{BIT360} & \multicolumn{2}{c}{VQA-ODV}\\
 &  SRCC  & PLCC & SRCC & PLCC \\
 \hline
S-PSNR     &    0.65    &        0.47  & 0.70 &   0.69 \\
Li18  & - & - & 0.80 & 0.78 \\
Li18-human  & - & - & \textbf{0.83} & 0.81\\
\hline
MC360IQA      &   0.67   &  0.75    &  0.30 &  0.34   \\
ERP-VQA     &    0.90   &    0.93   & 0.78  & 0.78 \\
Scanpath-VQA  & - & - & 0.81 & \textbf{0.83} \\
\hline
Proposed      &   \textbf{0.94}   &   \textbf{0.94}  &  0.80 &  0.79  \\ 
\whline
\end{tabular}

\end{table}

\begin{table}[t]
\caption{Inference time in seconds of our method against its two variants on a 15-second VR video with a resolution of $3,840 \times 1,920$ and a frame rate of $30$ fps}
\label{table:inference_time}
\renewcommand\arraystretch{1.25}
\centering
\begin{tabular}{l|c}
\whline
Model       & Inference Time \\ \hline

ERP-VQA      & 4.09                     \\
Scanpath-VQA & 173.77                   \\
\hline
Proposed        & 4.14                     \\ 
\whline
\end{tabular}
\end{table}

Additionally, we train and test the proposed method on  two more datasets BIT360~\cite{Zhang2018_subjective} and VQA-ODV \cite{li2018_subjective}, encompassing synthetic distortions such as compression, downsampling, and projection. As pristine-quality reference videos are available, we include a knowledge-driven full-reference VQA (FR-VQA) model, S-PSNR~\cite{Yu2015SPSNR_OIQA}, and a data driven FR-VQA model~\cite{li2018_subjective} that leverages predicted~\cite{xu2018predicting} and human scanpaths, respectively, denoted by Li18 and Li18-human. We also implement ERP-VQA and Scanpath-VQA, and incorporate a recent BIQA model, MC360IQA~\cite{Sun2020MC360IQA} for comparison. 
The database splitting strategies follow the descriptions of the original papers, ensuring content independence during training and testing. Table~\ref{table:qa_performance_bit360_vqaodv} lists the results, where we find that  our model performs the best on BIT360. Regarding VQA-ODV, our method is comparable with the best performers Li18-human and Scanpath-VQA, which necessitates human scanpaths for prediction.  These results  highlight the generality of our proposed model for assessing the perceptual VR video quality.

We further compare the inference time of our method against its two variants: ERP-VQA and Scanpath-VQA in Table~\ref{table:inference_time}. It is clear that the added computational complexity of pseudocylindrical convolution over standard convolution is marginal. In contrast, Scanpath-VQA needs to evaluate and average the visual quality of multiple planar videos from different scanpaths, which is the slowest.

\section{Conclusion and Discussion}
We have put together VRVQW, the first in the wild VR video database that includes MOSs and viewing behavioral data. We conducted a psychophysical experiment on VRVQW, involving $139$ users viewing VR videos from two conditions, resulting in a total of $40,268$ opinion scores and scanpaths. We performed a statistical analysis of various effects on the viewing behaviors and perceived quality in VR. Furthermore, we proposed the first blind VQA model for VR videos, which performs  favorably on VRVQW and two  VR video compression databases.

While our work presents an initial effort to understand the perceived quality of VR videos in the wild, there are still many important research problems that remain unexplored. 
First, in our psychophysical experiment, we manually minimize the adverse physiological reactions by means of questionnaires. It would be interesting to design ingenious psychophysical experiments to disentangle visual discomfort and visual quality in a quantifiable way. 
Second, when collecting viewing behaviors, users are exposed to a VR scene only once in order to eliminate any prior knowledge of the scene configuration. However, it would be valuable to investigate how viewing behaviors change with repeated exposures to the same scene and how these changes impact the perceived quality of the scene. 
Third, objective quality models tailored to VR videos in the wild are largely lacking. Our study suggests that a key step  is to incorporate viewing conditions. Scanpath-VQA that faithfully reflects how humans explore VR videos leads to improved performance. However, to fully automate Scanpath-VQA, scanpath prediction is inevitable, which is yet another challenging computational vision problem.
Currently,  the prediction of scanpaths for VR videos, particularly for a long-term horizon (e.g., $\ge10$ seconds), is still a nascent area of research. Therefore, there is a need to develop more advanced scanpath prediction models that can provide accurate short-term and long-term predictions while also capturing the diverse nature of human scanpaths. 
Last but not least, the current work considers VR videos as the sole
visual stimuli, it is of interest to investigate audio-visual perception
of VR videos \cite{Chao2020_Audio-Visual} and its consequences on the perceived quality.

\section{Acknowledgement}
This work was supported in part by National Natural Science Foundation of China (NSFC) under Grants 62102339 and 62132006, Shenzhen Science and Technology Program, PR China under Grant RCBS20221008093121052 and GXWD20220811170130002, and the Hong Kong RGC Early Career Scheme (9048212).

\bibliographystyle{IEEEtran}
\bibliography{bib}



\begin{IEEEbiography}[
{\includegraphics[width=1.5in,height=1.3in,clip,keepaspectratio]{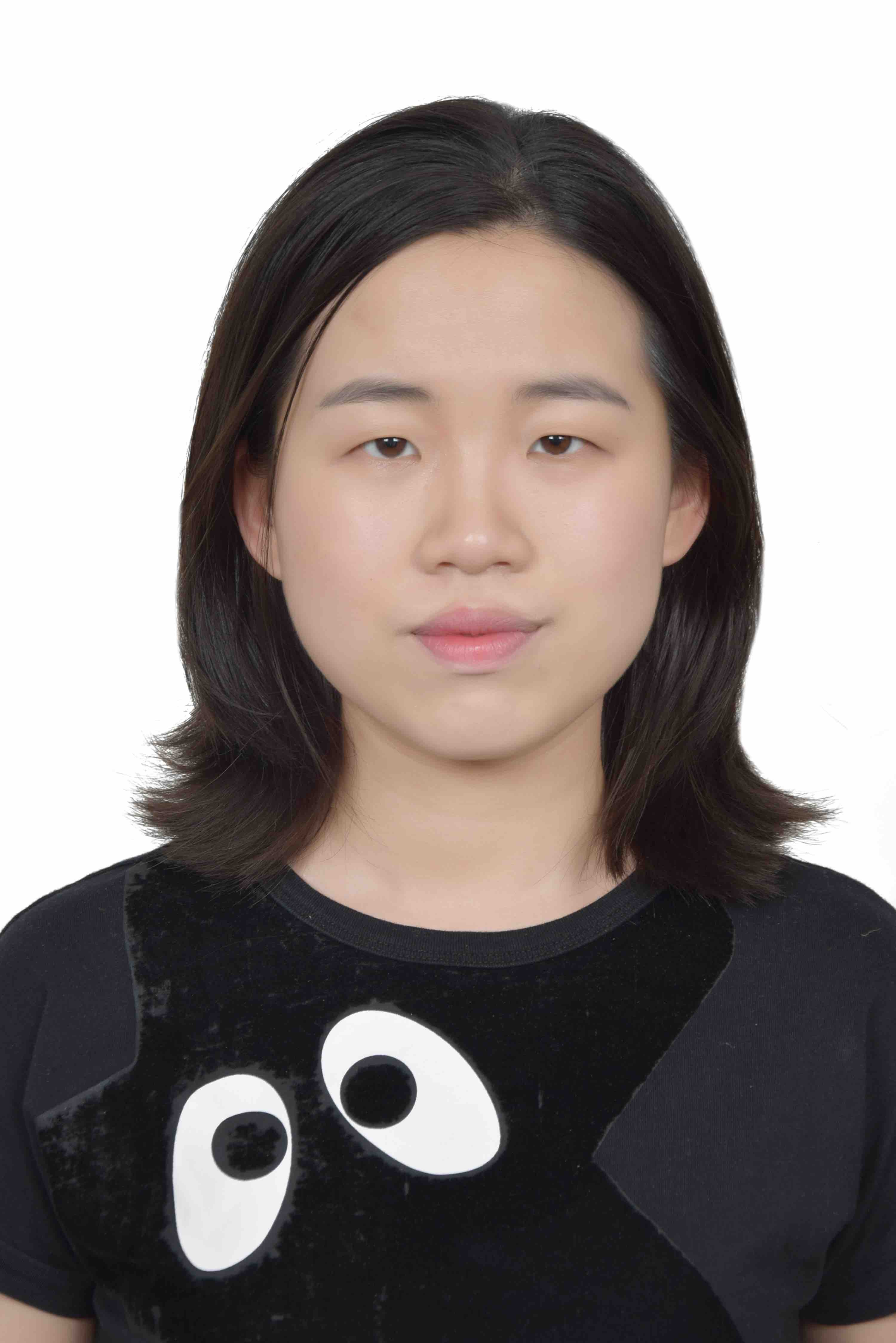}}]{Wen Wen} received the B.Sc. and M.A.Sc. degrees from Sun Yat-sen University, Guangzhou, China, in 2019 and 2021, respectively. She is currently pursuing the Ph.D. degree with the Department of Computer Science, City University of Hong Kong, Kowloon, Hong Kong. Her research interests include quality assessment and virtual reality.
\end{IEEEbiography}\vspace{-4em}

\begin{IEEEbiography}[
{\includegraphics[width=1.5in,height=1.3in,clip,keepaspectratio]{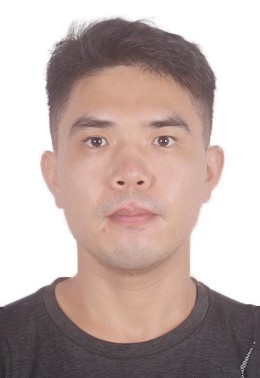}}]
{Mu Li} received his B.E. degree in computer science and technology in 2015 from Harbin Institute of Technology, and the Ph.D. degree from the Department of Computing, the Hong Kong Polytechnic University, Hong Kong, China, in 2020. He worked as a Postdoctoral Researcher at the Chinese University of Hong Kong, Shenzhen.  He is currently with Harbin Institute of Technology, Shenzhen, China. His research interests include deep learning, image processing, image compression, and virtual reality.
\end{IEEEbiography}\vspace{-4em}

\begin{IEEEbiography}[
{\includegraphics[width=1.5in,height=1.3in,clip,keepaspectratio]{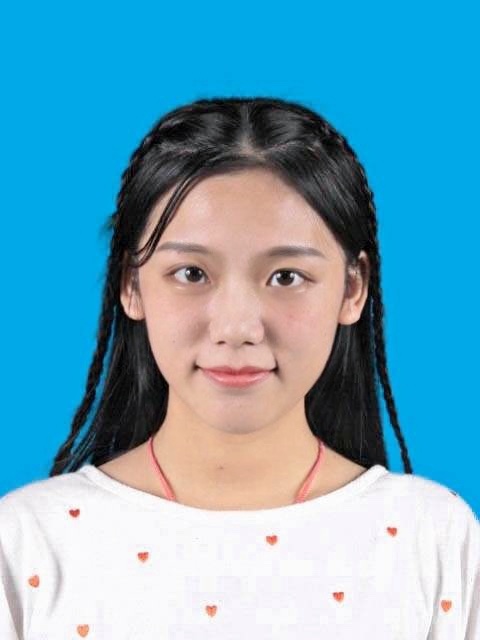}}]
{Yiru Yao} received the B.E. and M.A.Sc. degrees from the Jiangxi University of Finance and Economics, Nanchang, China, in 2020 and 2023, respectively. Her research interests include visual quality assessment and VR image/video processing.
\end{IEEEbiography}\vspace{-4em}

\begin{IEEEbiography}[
{\includegraphics[width=1.5in,height=1.3in,clip,keepaspectratio]{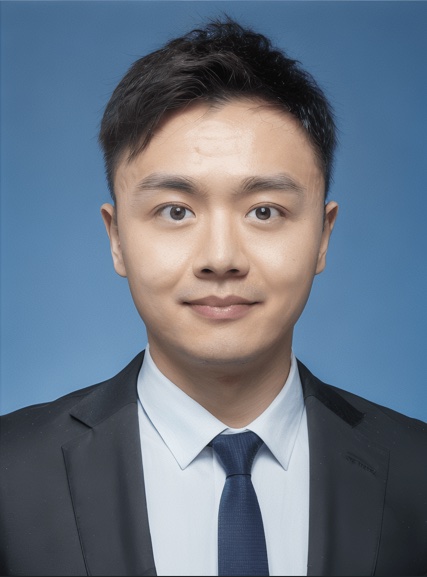}}]
{Xiangjie Sui} received the B.E. and M.A.Sc. degrees from the Jiangxi University of Finance and Economics, Nanchang, China, in 2018 and 2021, respectively, where he is currently pursuing the Ph.D. degree. His research interests include visual quality assessment and VR image/video processing.
\end{IEEEbiography}\vspace{-4em}

\begin{IEEEbiography}[
{\includegraphics[width=1.5in,height=1.3in,clip,keepaspectratio]{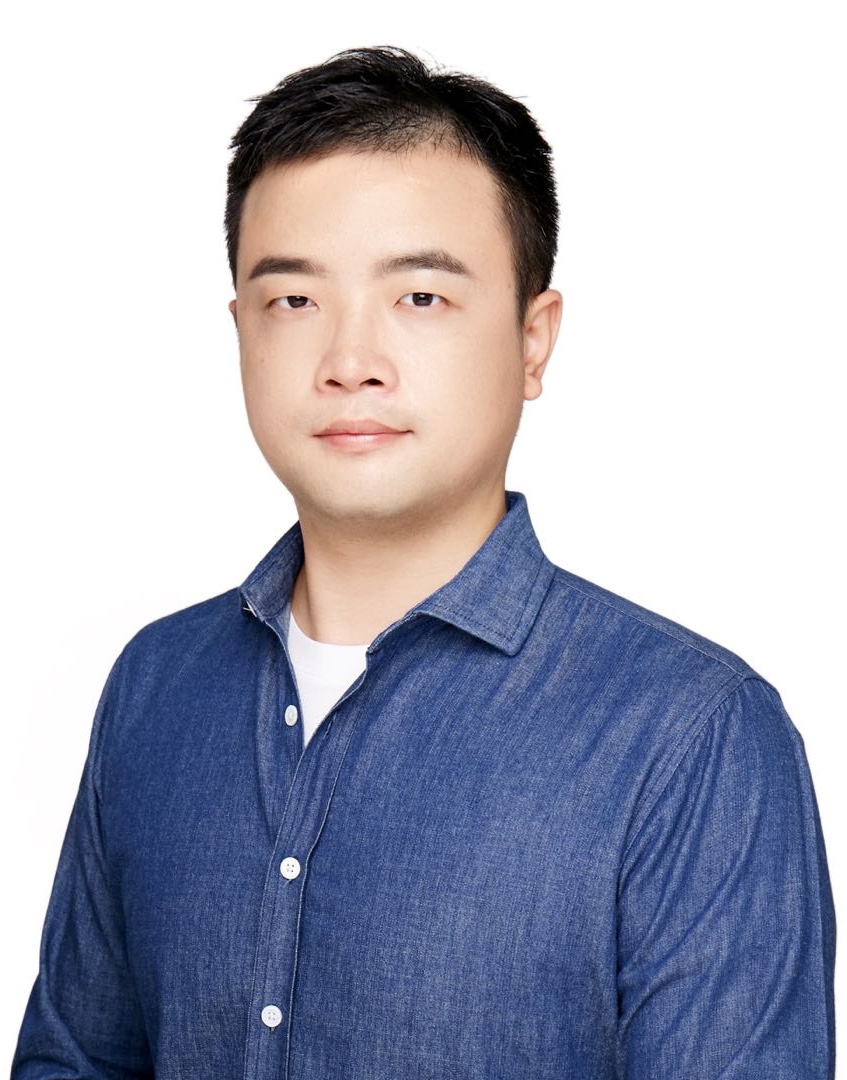}}]
{Yabin Zhang} received the B.E. degree in electronic information engineering in the Honors School, Harbin Institute of Technology and the Ph.D. degree from the School of Computer Science and Engineering, Nanyang Technological University, Singapore in 2013 and 2018, respectively. He is currently a Principle Researcher in Multimedia Lab, ByteDance, Shenzhen.  His research interests include video coding, image/video processing, image/video quality assessment, and computer vision.
\end{IEEEbiography}\vspace{-4em}

\begin{IEEEbiography}[
{\includegraphics[width=1.5in,height=1.3in,clip,keepaspectratio]{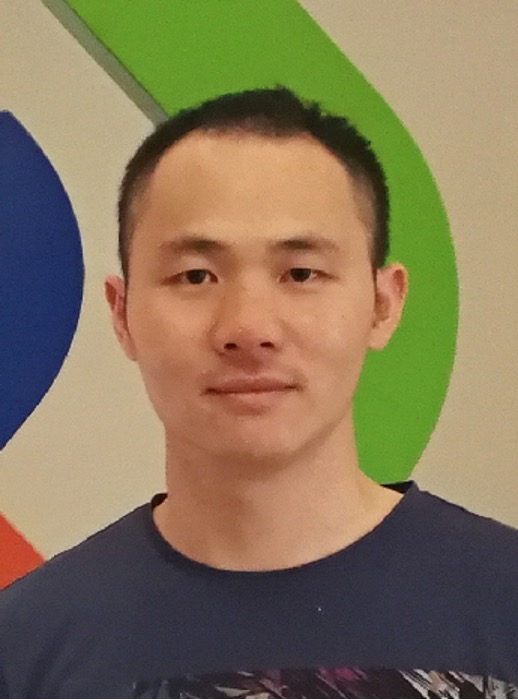}}]
{Long Lan} is an Associate Professor of College of Computer Science and Technology at National University of Defense Technology. He was a Visiting PhD Student at University of Technology, Sydney (2015-2017). He received his Ph.D. degree in Computer Science from National University of Defense Technology (2013- 2017). His research interests lie in computer vision and machine learning, especially multi-object tracking, transfer learning and few shot learning. He has served as the senior program committee of IJCAI’21, and published 40+ journal articles and conference papers, such as IJCV, TIP, TMM, TKDE, TCSVT, NeurIPS, ICLR, MM, KDD, AAAI, IJCAI, and ECCV.
\end{IEEEbiography}\vspace{-4em}

\begin{IEEEbiography}[
{\includegraphics[width=1.5in,height=1.3in,clip,keepaspectratio]{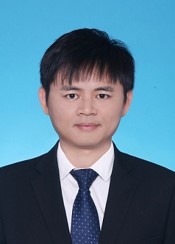}}]
{Yuming Fang} (Senior Member, IEEE) received the B.E. degree from Sichuan University, Chengdu, China, the M.S. degree from the Beijing University of Technology, Beijing, China, and the Ph.D. degree from Nanyang Technological University, Singapore. He is currently a Professor with the School of Information Management, Jiangxi University of Finance and Economics, Nanchang, China. His research interests include visual attention modeling, visual quality assessment, computer vision, and 3D image/video processing. He serves on the editorial board for IEEE Transactions on Multimedia and Signal Processing: Image Communication.
\end{IEEEbiography}\vspace{-4em}

\begin{IEEEbiography}[
{\includegraphics[width=1.5in,height=1.3in,clip,keepaspectratio]{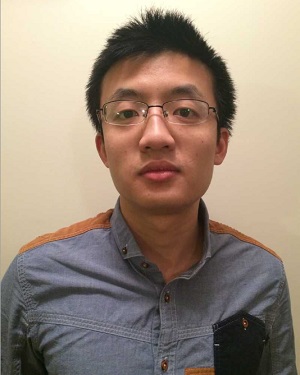}}]
{Kede Ma} (Senior Member, IEEE) received the B.E. degree from the University of Science and Technology of China, Hefei, China, in 2012, and the M.S. and Ph.D. degrees in electrical and computer engineering from the University of Waterloo, Waterloo, ON, Canada, in 2014 and 2017, respectively. He was a Research Associate with the Howard Hughes Medical Institute and New York University, New York, NY, USA, in 2018. He is currently an Assistant Professor with the Department of Computer Science, City University of Hong Kong. His research interests include perceptual image processing, computational vision, computational photography, and multimedia forensics and security.
\end{IEEEbiography}\vspace{-4em}

\end{document}